\documentclass[10pt,twocolumn,letterpaper]{article}

\usepackage{wacv}
\usepackage{times}
\usepackage{epsfig}
\usepackage{graphicx}
\usepackage{amsmath}
\usepackage{amssymb}
\usepackage{booktabs}
\usepackage{float} 
\usepackage{caption}
\usepackage{subcaption}
\usepackage{comment}
\usepackage{amsmath,amssymb}
\usepackage{color}
\usepackage{subcaption}
\usepackage{booktabs}
\usepackage{multirow}
\usepackage{multicol}
\usepackage{times}
\usepackage{epsfig}
\usepackage{graphicx}
\usepackage{enumitem}
\usepackage{subcaption}
\usepackage{pifont}
\usepackage{dsfont}
\usepackage{wrapfig} 
\usepackage{float} 
\usepackage[normalem]{ulem}
\usepackage{graphbox}

\makeatletter
\@namedef{ver@everyshi.sty}{}
\makeatother
\usepackage{tikz}

\makeatletter
\renewcommand\paragraph{%
  \@startsection{paragraph}{4}{\z@}%
     {0.8ex \@plus3ex \@minus.2ex}%
     {-0.5em}%
     {\normalfont\normalsize\bfseries}}
\makeatother

\def\wacvPaperID{282} % Enter the WACV Paper ID here
\wacvfinalcopy %

\ifwacvfinal
\usepackage[breaklinks=true,bookmarks=false]{hyperref}
\else
\usepackage[pagebackref=true,breaklinks=true,colorlinks,bookmarks=false]{hyperref}
\fi

\usepackage{color}

\newif\ifshowedits

\newcommand{\addeditor}[3]{%
  \definecolor{#1color}{rgb}{#3}
  \expandafter\newcommand\csname #1\endcsname[1]{%
  \ifshowedits
    {\color{#1color} ##1}%
  \else
    {##1}%
  \fi
  }%
  \expandafter\newcommand\csname #1rmk\endcsname[1]{%
  \ifshowedits
    {\color{#1color} {\bf [#2: ##1]}}
  \fi
  }%
  \expandafter\newcommand\csname #1rpl\endcsname[2]{%
  \ifshowedits
    {\color{#1color} ##1 \sout{##2}}
  \else
    {##1}
  \fi
  }%
}

\newcommand{\mycomment}[1]{}

\newcommand{\calA}{{\cal A}}

\newcommand{\calD}{{\cal D}}

\newcommand{\calF}{{\cal F}}

\newcommand{\calJ}{{\cal J}}

\newcommand{\calL}{{\cal L}}

\newcommand{\calW}{{\cal W}}

\newcommand{\bd}{{\bf d}}

\newcommand{\bx}{{\bf x}}

\newcommand{\bX}{{\bf X}}

\newcommand{\bXtwo}{{\bX_2}}

\newcommand{\IR}{{\mathds{R}}}

\DeclareMathOperator*{\argmax}{arg\,max}

\DeclareFontFamily{U}{mathx}{\hyphenchar\font45}
\DeclareFontShape{U}{mathx}{m}{n}{
      <5> <6> <7> <8> <9> <10>
      <10.95> <12> <14.4> <17.28> <20.74> <24.88>
      mathx10
      }{}
\DeclareSymbolFont{mathx}{U}{mathx}{m}{n}
\DeclareFontSubstitution{U}{mathx}{m}{n}
\DeclareMathAccent{\widebar}{0}{mathx}{"73}

\newcommand{\dino}[2]{\psi_{#1}^{#2}}
\newcommand{\flowsymb}{\phi}
\newcommand{\flow}[3]{\flowsymb_{#1,#2}^{#3}}

\newcommand{\mask}[1]{\bx_{#1}}
\newcommand{\maskini}[1]{\widehat{\bx}_{#1}}

\newcommand{\warp}[2]{w_{#1}^{#2}}

\newcommand{\aff}[1]{A_{#1}}
\newcommand{\flw}[1]{F_{#1}}
\newcommand{\overbar}[1]{\mkern 1.5mu\overline{\mkern-1.5mu#1\mkern-1.5mu}\mkern 1.5mu}
\newcommand{\rflw}[1]{\overbar{F\,}\!\!_{#1}}

\newcommand{\inv}{{\;\text{--}1}}

\newcommand{\sdotprod}[2]{\langle#1,#2\rangle}

\newcommand{\CE}{\text{CE}}
\newcommand{\nint}[1]{\ensuremath\left\lfloor#1\right\rceil}

\newcommand{\cmark}{\ding{51}}%
\newcommand{\Nsteps}{T}

\addeditor{vincent}{VL}{0.0, 0.5, 0.0}
\addeditor{renaud}{RM}{1.0, 0.5, 0}
\addeditor{georgy}{GP}{0.0, 0.0, 0.5}
\addeditor{dym}{DYM}{1.0, 0.0, 0.0}
\addeditor{yang}{YX}{0.3, 0.5, 0.7}
\addeditor{nermin}{NS}{0.5, 0.5, 0.0}
\showeditsfalse

\begin{document}

%%%%%%%%% TITLE
\title{
A Simple and Powerful Global Optimization for\\
Unsupervised Video Object Segmentation
\vspace{-0.5cm}
}

\renewcommand{\and}{\hspace{0.25cm}}

\author{
\hspace{-5mm}
Georgy Ponimatkin$^1$
\and 
Nermin Samet$^1$
\and
Yang Xiao$^1$
\and
Yuming Du$^1$
\and
Renaud Marlet$^{1,2}$
\and
Vincent Lepetit$^{1}$
\\[2mm]
% \and
\hspace{-5mm}\textsuperscript{1}LIGM, Ecole des Ponts, Univ Gustave Eiffel, CNRS, Marne-la-Vall\'ee, France
\hspace{2.5mm}\textsuperscript{2}Valeo.ai, Paris, France
\\
\small{\texttt{georgy.ponimatkin@enpc.fr}}\\
\small{Code and supplementary material:  \url{https://ponimatkin.github.io/ssl-vos}}
}

\maketitle

\begin{abstract}
    We propose \renaud{a simple, yet powerful} approach for unsupervised object segmentation in videos. We introduce an objective function whose minimum represents the mask of the main salient object over the input sequence. \renaud{It only relies on independent image features and optical flows, which can be obtained using off-the-shelf self-supervised methods.} It scales with the length of the sequence with no need for superpixels or sparsification, and it generalizes to different datasets \renaud{without any specific training}. This objective function can \renaud{actually} be derived from \renaud{a form} of spectral clustering applied to the entire video.  \renaud{Our} method achieves on-par performance with the state of the art on \renaud{standard benchmarks} (DAVIS2016, SegTrack-v2, FBMS59), while being conceptually \renaud{and practically} much \renaud{simpler}. 

% \url{https://ponimatkin.github.io/ssl-vos}.

\end{abstract}

\section{Introduction}
\label{sec:introduction}

While the two research communities working on unsupervised video object segmentation~\cite{mahadevan2020making,athar2020stem,yang2019anchor,lu2019see} and on object discovery~\cite{vo2019unsupervised,yang2021self,Lamdouar21,LOST} often remain separated, they share the goal of segmenting objects in visual data without depending on manual labels for these objects. This ability is essential to autonomous systems for evolving and interacting in open world. It is also a fundamental problem for Computer Vision as humans have the ability to learn about new objects without guidance.

Object appearance and motion are important cues to achieve this task. However, many challenges remain: Objects can share similar appearance with the background, their visual appearance may not be uniform, and different parts can move in different directions. As a result, many methods still rely on some sort of supervision, at least for learning to extract visual features.

\begin{figure}
\begin{subfigure}{0.32\columnwidth}
  \centering
  % include first image
  \includegraphics[width=1\textwidth,trim={8cm 6cm 5cm 2cm},clip]{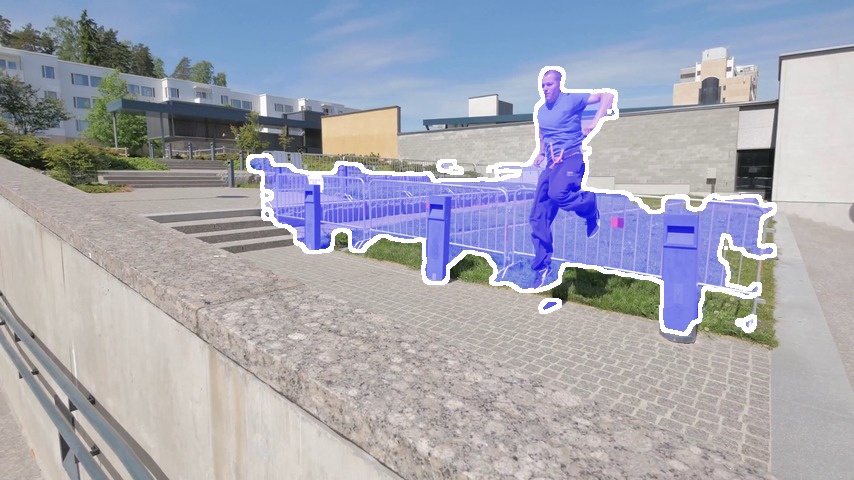}
  \vspace*{-6mm}
  \caption{}
\end{subfigure}
\begin{subfigure}{0.32\columnwidth}
  \centering
  % include second image
   \includegraphics[width=1\textwidth,trim={8cm 6cm 5cm 2cm},clip]{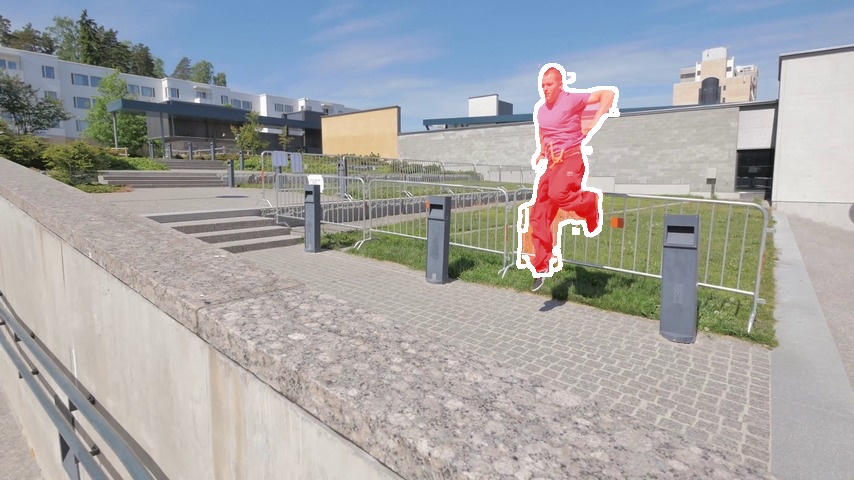} 
  \vspace*{-6mm}
  \caption{}
  \label{fig:sub-second}
\end{subfigure}
\begin{subfigure}{0.32\columnwidth}
  \centering
  % include second image
  \includegraphics[width=1\textwidth,trim={8cm 6cm 5cm 2cm},clip]{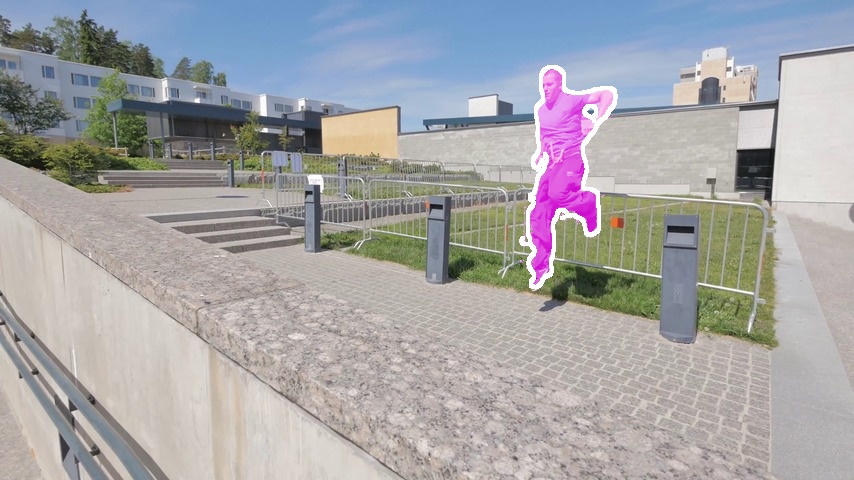}
  \vspace*{-6mm}
  \caption{}
  \label{fig:sub-second}
\end{subfigure}
\vspace{-3mm}
\caption{ \textbf{(a)} Segmentation obtained by our spectral clustering formulation on the self-supervised image features from DINO~\cite{caron2021emerging} in a single frame. \textbf{(b)} Segmentation obtained using the same self-supervised image features \emph{and} optical flow from ARFlow~\cite{liu2020learning}, but still from a single frame. \textbf{(c)} Final segmentation obtained with our complete method, after optimization on the full frame sequence, using the same features and optical flow estimated by ARFlow. }
\label{fig:teaser}
\vspace{-2mm}
\end{figure}

We present here a simple approach to unsupervised video object segmentation. It leverages recent progress on unsupervised learning in a simple but powerful, novel optimization scheme over the objects' masks. We start from self-supervised features such as DINO~\cite{caron2021emerging}, MoCo-v3~\cite{chen2021mocov3}, SWAV~\cite{caron2020unsupervised} or Barlow Twins~\cite{zbontar2021barlow}  as the appearance cue, and the optical flow from methods such as RAFT~\cite{teed-eccv20-raft} or ARFlow~\cite{liu2020learning}. DINO, MoCo-v3, SWAV, Barlow Twins and ARFlow methods are ``self-supervised'' or ``unsupervised'', in the sense that they do not use any manual annotations, making our method also entirely unsupervised.\footnote{The terminology is still fluctuating, as the DINO method is called ``self-supervised'' in the original paper, while ARFlow is called ``unsupervised''. Throughout the paper, we will use self-supervised and unsupervised in the sense of ``not using manual annotations''.}

Used alone, the DINO features can already provide a surprisingly good segmentation of the objects, but still below the state of the art\renaud{, in particular for videos. Yet,} with our optimization scheme, we can reach a performance that is on par or even better than much more sophisticated methods.

\vincent{
Our optimization starts from an initial estimate for the object's mask in each frame of the input sequence. It then optimizes the masks over the whole video sequence. We show that our optimization function can be derived from spectral clustering over the video sequence. \renaud{Yet,} spectral clustering is difficult to make tractable for long sequences \renaud{as} 
it requires computing one of the eigenvectors of a huge matrix~\cite{Shi00,meila}. To obtain a tractable problem, previous methods \renaud{based on spectral clustering} rely on superpixels~\cite{fabio2012video} or graph sparsification~\cite{khoreva2015classifier}. 
However, superpixels may introduce artefacts at object boundaries, 
\renaud{and both superpixels and sparsification require careful parameter tuning.}
\renaud{Besides, even using fast methods to extract relevant eigenvectors, such as Power Iteration Clustering~(PIC)~\cite{power-iteration}, the complexity of these formulations is \emph{a priori} quadratic in the length of the sequence (depending however on sparsity).} 
In contrast, our \renaud{method only needs eigenvectors computed for each frame independently, based on image features and optical flow for the frame, thus essentially scaling linearly with the number of frames.}
In short, our approach yields a tractable approximation of spectral clustering over the entire sequence. \renaud{Concretely, on videos from the DAVIS2016 dataset~\cite{perazzi-cvpr16-abenchmarkdataset}, our method is on average} 
\nermin{$\sim$170$\times$ faster than our full spectral clustering
counterpart, \ie, TokenCut~\cite{wang2022tokencut}.}
}

\begin{figure}
  \centering
  \begin{tabular}{ccc}
    \includegraphics[width=0.2\linewidth,trim={7cm 2cm 22cm 3cm},clip]{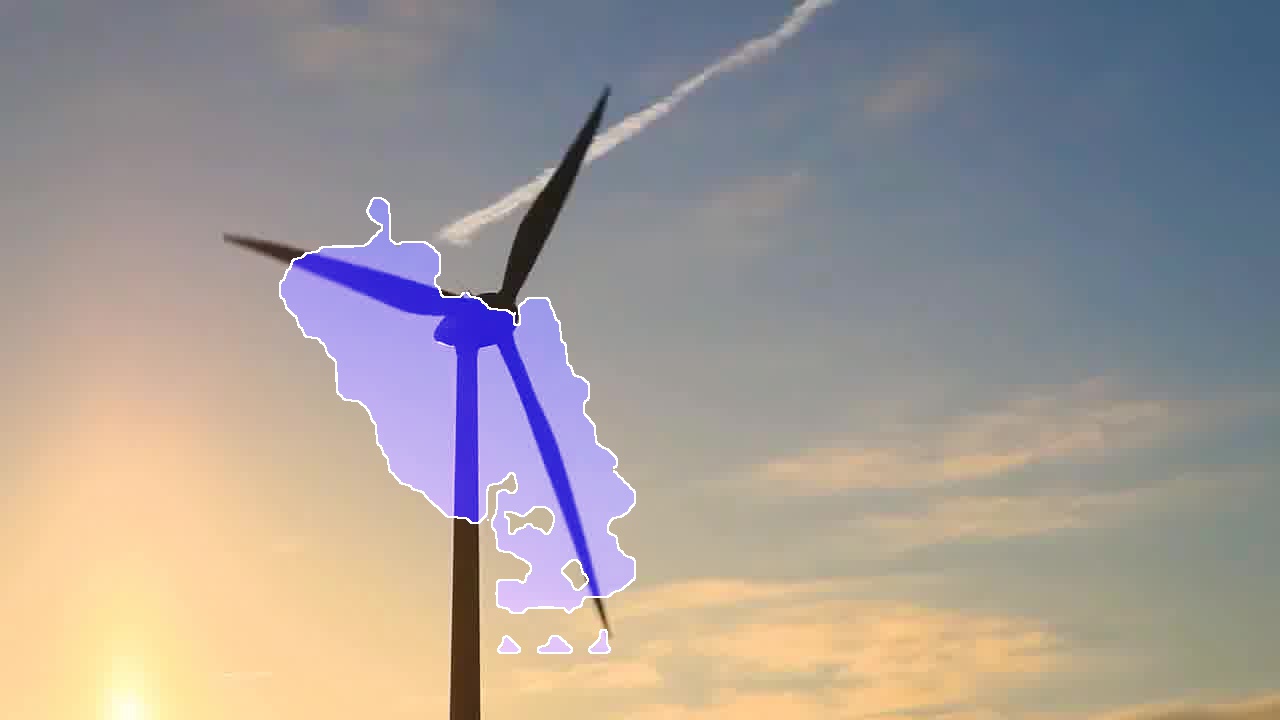} &
    \includegraphics[width=0.2\linewidth,trim={7cm 2cm 22cm 3cm},clip]{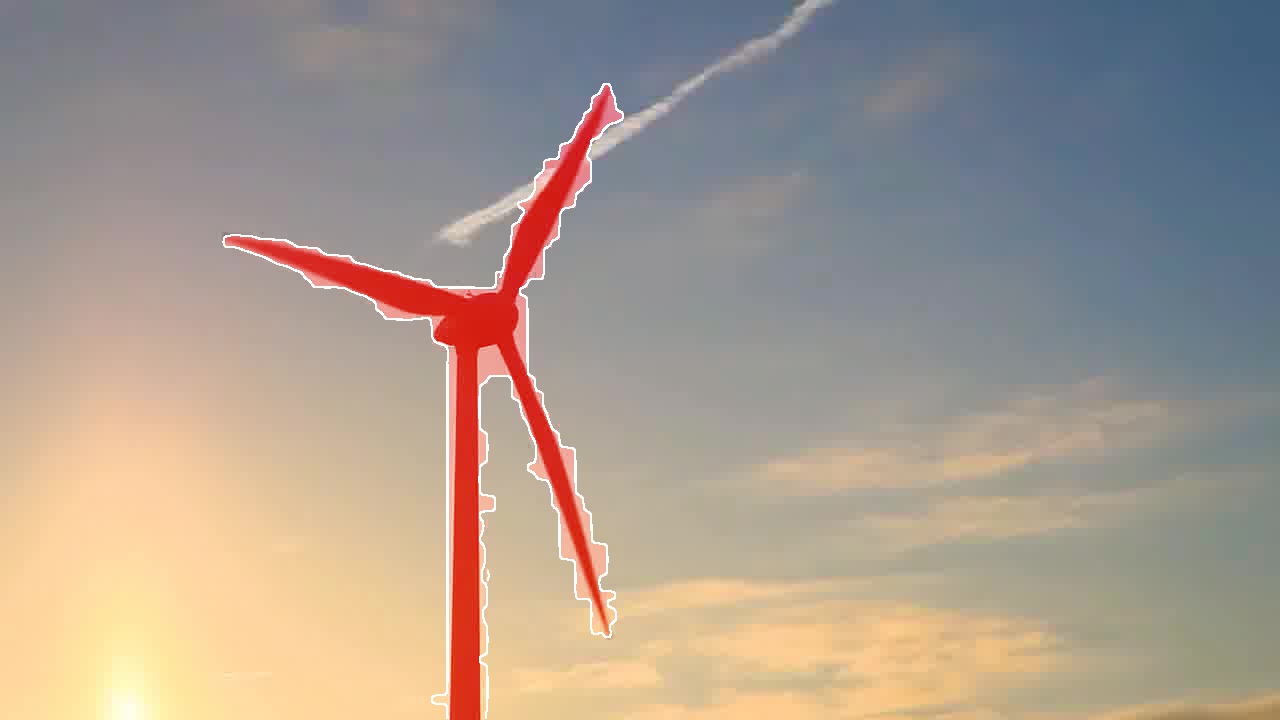} &
    \includegraphics[width=0.2\linewidth,trim={7cm 2cm 22cm 3cm},clip]{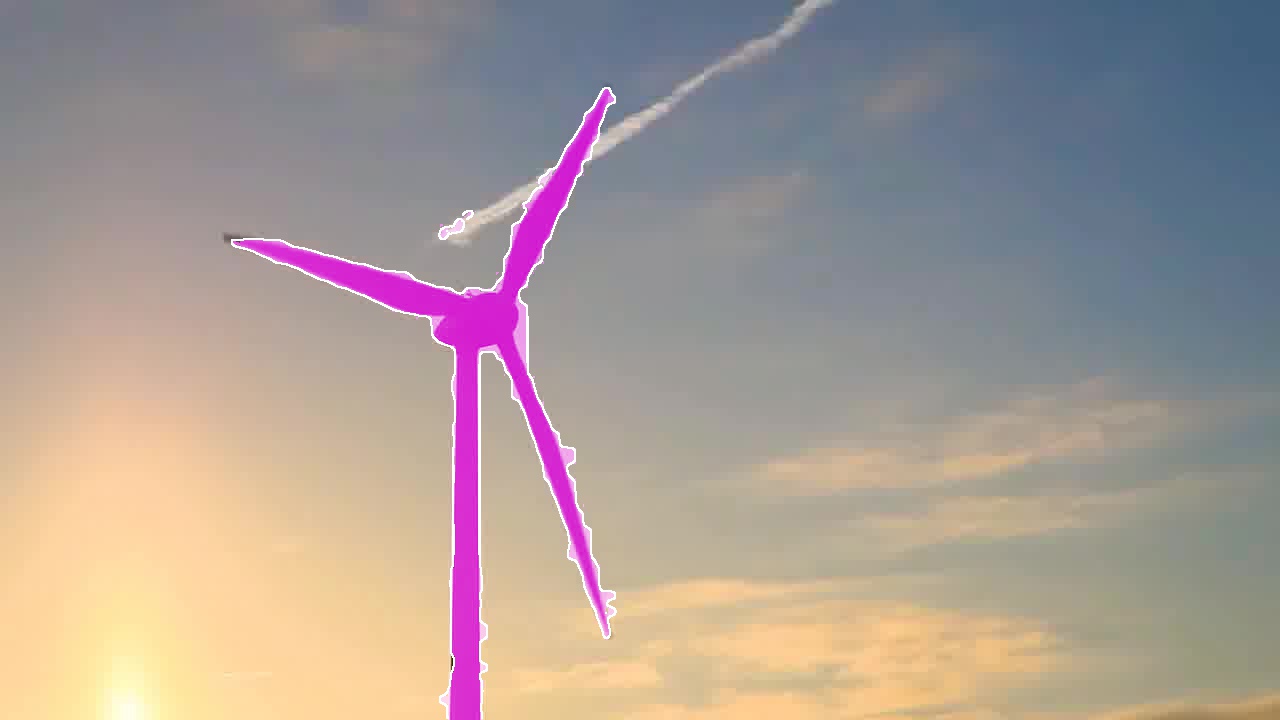} \\
    (a) & (b) & (c)\\
  \end{tabular}
  \vspace*{-2mm}
  \caption{\label{fig:clustering} \textbf{Generalization to new datasets. } Relying on spectral clustering rather than learning generalizes better to new datasets. (a) The learned method of \cite{yang2021self} performs poorly on a new sequence.  (b) Spectral clustering on the \emph{same} optical flow performs significantly better, even without optimization. (c) Result of our complete method.}
  \vspace*{-2mm}
\end{figure}

Importantly, for practical applications, our method can be applied to different datasets without retraining. This is in contrast with \renaud{learning-based segmentation approaches.}
As illustrated in Fig.\,\ref{fig:clustering}, they do not generalize well to new data.

To summarize, our contribution is the derivation of a novel objective function from spectral clustering, which results in a simple and efficient optimization  method for object discovery and segmentation in videos.  \renaud{Moreover,} our method is \renaud{arguably} the first that relies purely on self-supervised features, 
without the need of any manual annotation. Still, it outperforms previous methods on several standard challenging datasets, \renaud{including}
some methods that only rely on supervised training. Last, it can \renaud{directly} be
applied to different datasets, without \renaud{any} training or tuning.

\section{Related Work}
\label{sec:relatedwork}

We discuss here works on object discovery and unsupervised video object segmentation, as both are closely related to our work. We also discuss
works on unsupervised feature learning, which
we rely on
for image feature extraction.

\subsection{Object Discovery}

Object discovery aims to localize the objects in images or videos. Some works rely on a collection of images containing objects of the same class and localize these objects using clustering~\cite{faktor2013clustering}, image matching~\cite{rubinstein2013unsupervised,cho2015unsupervised}, topic discovery \cite{russell2006using,sivic2008unsupervised} or optimization for selecting region proposals~\cite{vo2019unsupervised}. One limitation of these works is that \renaud{such a collection could be difficult to obtain, and its quality would heavily impact performance.}
Du \etal~\cite{Du_2021_ICCV} proposes to discover and segment objects from unseen classes based on instance masks predicted by models trained only on seen classes, while our method does not need any supervision.

Recently, some works~\cite{locatello2020object,burgess2019monet,greff2019multi} adopt a bottom-up approach and segment the object in the images by exploiting the similarity among pixel colors or patch features. These works have only demonstrated their performance on synthetic images and can be easily affected by image texture or colors. In contrast, our approach aims to discover objects using unlabeled in-the-wild videos without any supervision. 

\cite{yang2021self,Lamdouar21} and \cite{haller2021iterative} can also be considered as object discovery methods, as their goal is to segment the primary objects given some video sequences. \cite{yang2021self,Lamdouar21} propose to use an attention-based network to compute the trajectory embeddings from optical flow only. One major limitation of \cite{yang2021self,Lamdouar21} is that optical flow is not sufficient to segment static objects. Besides, even though \cite{yang2021self,Lamdouar21} do not require any labeled data, they still have to train their network with different strategies on different datasets, which makes it difficult to generalize, as show in Figure \ref{fig:clustering}. As we will show in Section \ref{sec:experiments}, our method does not need any training and can achieve good results on all the benchmarks.

IKE~\cite{haller2021iterative} proposes an iterative refinement approach: In the first stage, the videos are first fed into a graph module for mask initialization. Then, in the second stage, the initialized masks are used to train a segmentation network, whose predictions are used as initialization for the graph module in the first stage. In the first stage, optical flow is used to generate long-term space-time trajectories. Optical flow can be very noisy and thus it may be difficult to obtain long trajectories reliably in general. We will show in the Experiments section that our method can achieve better results compared to \cite{haller2021iterative} while at the same time being simple. 
Some methods~\cite{yang2021self,Lamdouar21,yang2019unsupervised} are also presented as self-supervised or unsupervised while they \renaud{actually} rely on supervised optical flow methods such as RAFT~\cite{teed-eccv20-raft}. In comparison, our method is able to achieve state-of-the-art performance while \renaud{only} using self-supervised optical flow learned from videos.

\subsection{Unsupervised Video Object Segmentation} 

Given a video sequence, Unsupervised Video Object Segmentation~(UVOS) aims to consistently segment and track the most salient objects in the video without any human intervention. The approaches for UVOS can be divided into two categories depending on whether they need labeled data or not. 
Methods such as \cite{mahadevan2020making,yang2019anchor,lu2019see,athar2020stem,wang2019learning,oh2019video,li2018video,voigtlaender2017online,maninis2018video,perazzi2017learning,wang2019zero,song2018pyramid} use either labeled images or labeled videos to train their networks for video segmentation. In addition, some works \cite{zhou2020motion,xie2019object,yang2021dystab,ren2021reciprocal,zhang2021deep,yang2021learning,ji2021full,tokmakov2017learning,cheng2017segflow,siam2019video,jain2017fusionseg,tokmakov2017learning} also exploit low-level cues such as object boundaries to get better mask predictions. One main limitation of these methods is that they rely heavily on their large-scale well-annotated training data, which can be hard to get in practice. In contrast, our proposed method is entirely self-supervised and does not need any labeled data. 

There are also several works~\cite{Faktor2014VideoSB,croitoru2017unsupervised,papazoglou2013fast,bideau2016s,brox2010object,ranjan2019competitive,keuper2015motion,ochs2011object,sivic2004object,yang2019unsupervised,wang2015saliency,yang2021self,Ye2022CVPR} that can segment the primary objects in the video without any labeled data. One major limitation of all these works is that they all consider that the pixels on the objects in the video share similar motion patterns, thus, their methods can fail when the objects are static or move with the same speed as the background, while our proposed method combines the appearance cues and the motion cues and can largely alleviate this issue.  

Currently, DyStaB~\cite{yang2021dystab} is the best performing unsupervised method for video object segmentation. The method consists of three parts: A static model and a dynamic model~(both based on DeepLab backend~\cite{chen2017deeplab}), together with an inpainter network based on~\cite{yang2019unsupervised}. The three networks are jointly trained via an adversarial loss in an iterative fashion to obtain final segmentation results and and significant post-processing via CRF is utilized. In comparison, our approach does not require training---besides for the image features and optical flow that can be obtained out-of-the-shelf---while an adversarial loss can be difficult to train. More importantly, our approach is much simpler, while achieving a similar performance to DyStaB.

\subsection{Self-Supervised Learning}

Recent self-supervised approaches~\cite{wu2018unsupervised,chen2020simple,dosovitskiy2014discriminative,he2020momentum,chen2021exploring,grill2020bootstrap,caron2021emerging} propose to train a feature extraction  network using an instance classification pipeline, which treats each image as a single class and trains the network to distinguish images cropped from a large image collection without any manual annotation. In particular, DINO~\cite{caron2021emerging} introduced an approach which makes a network trained in a self-supervised manner learn class-specific features. 

Motivated by this ability, discovering objects in images using self-supervised features recently gained attention~\cite{wang2022tokencut, melaskyriazi2022deep, LOST}. LOST~\cite{LOST} utilizes self-supervised features within a seed selection and expansion strategy and localizes the main object given an image. Closely related to LOST, TokenCut~\cite{wang2022tokencut} and Deep Spectral Methods~(DSM)~\cite{melaskyriazi2022deep} propose graph-based methods that use self-supervised transformer features to discover and segment salient objects. TokenCut~\cite{wang2022tokencut} builds a graph where visual tokens are nodes and similarity scores between tokens are edges of a weighted graph. They formulate the segmentation problem as a normalized graph cut and solve it using spectral clustering with eigendecomposition. Similar to TokenCut~\cite{wang2022tokencut}, inspired from traditional spectral segmentation methods, DSM~\cite{melaskyriazi2022deep} first constructs a Laplacian matrix which is a combination of color information and self-supervised transformer features. Next, the image is decomposed using the eigenvectors of the Laplacian matrix. 

Our method is similar to these methods as we also aim to detect and segment salient objects based on a graph formulation. There are two key differences: (i) We aim to discovery objects in videos rather than still frames and in order to ensure the temporal consistency, we extend the affinity matrix with optical flow to establish inter-frame connectivity; (ii) We introduce a novel method for optimizing on the entire video efficiently.
Moreover, we show that using power iteration clustering~\cite{power-iteration} instead of full eigenvector decomposition for spectral clustering as we do results in significantly faster run time (0.1s/frame vs. 17s/frame).

\section{Method}
\label{sec:method}
\begin{comment}
\georgy{
\paragraph{Goal.} Given a video sequence, our method produces a segmentation of the main salient object for each frame by optimizing a global objective function, and we first provide an overview of this objective function in Section~\ref{sec:overview}.

\paragraph{Approach.} The minimization of this objective function can be interpreted as a simplified and tractable variant of spectral clustering applied to the whole video sequence. In Section~\ref{sec:sc}, we briefly present spectral clustering and how it could be applied to video segmentation in principle.

\paragraph{Challenges.} This direct application is however intractable, and we show in Sections~\ref{sec:matrices} and \ref{sec:derivations} how it can be simplified to make it tractable, which finally yields our objective function.
}

\end{comment}

\setlength\intextsep{8mm} 
 \begin{figure*}[t]
 \centering
   \includegraphics[width=0.90\linewidth]{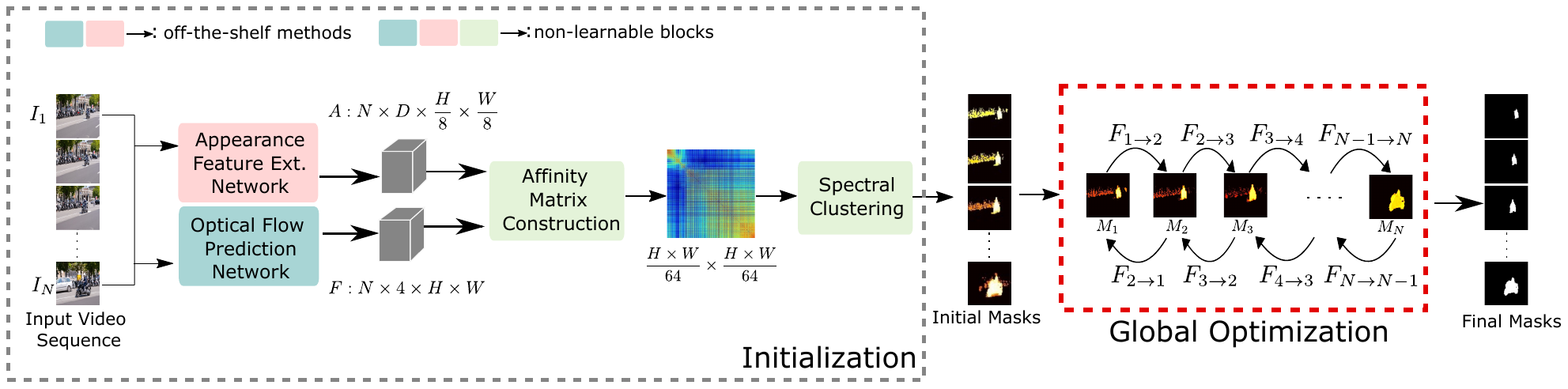}

  \caption{\label{fig:overview} \textbf{Overview of our approach. } Given a video sequence, starting from first estimates for the object masks obtained by spectral clustering on each image independently, we optimize the masks so that they remain close to the first estimates while being consistent with the optical flow. 
  The objective function we optimize to retrieve the masks can be derived from spectral clustering applied to the video sequence. Our method can rely on self-supervised visual features only.}
\end{figure*}

\subsection{Method Overview}
\label{sec:overview}

We consider a sequence of $N$ video frames of size $H\,{\times}\,W$, where $H$ and $W$ are the number of rows and columns of a frame. The segmentation of the main salient object, in each frame~$p$, is modeled as a soft, vectorized image mask $\mask{p} \,{\in}\, \IR_{+}^{HW}$. The actual image mask at frame $p$ is recovered by splitting the pixels into two clusters based on $\mask{p}$, separating the object of interest from the background.

We assume that we are given a rough mask estimate $\maskini{p}$ in each frame~$p$, as well as functions $\warp{p}{q}(\mask{p})$ warping a mask $\mask{p}$ in frame $p$ to frame $q$ using the optical flow from $p$ to $q$.  Then, the objective function we minimize is:
\begin{equation}
\begin{array}{l}
\calL(\{\mask{p}\}_p) = \sum_p \lambda \; \CE(\maskini{p}, \mask{p}) \; + \\[0.3em]
\quad \CE(\mask{p+1}, \warp{p}{p+1}(\mask{p}) ) + 
\CE(\mask{p}, \warp{p+1}{p}(\mask{p+1})) \> ,
\end{array}
\label{eq:obj}
\end{equation}
where $\CE(\cdot)$ denotes the cross-entropy.

Intuitively, $\CE(\maskini{p}, \mask{p})$ expresses the deviation between a mask $\mask{p}$ and the initial mask estimate $\maskini{p}$; $\CE(\mask{p+1}, \warp{p}{p+1}(\mask{p}))$ expresses the deviation between mask $\mask{p+1}$ and the warping of mask $\mask{p}$ by flow~$\warp{p}{p+1}$, and reciprocally for $\CE(\mask{p}, \warp{p+1}{p}(\mask{p+1}))$; finally, $\lambda$ is a constant weight to balance the significance of both kind of deviations (difference from initial mask vs flow discrepancy). 

As illustrated in Figure~\ref{fig:overview}, our objective function can be interpreted easily: The optimization starts from a first estimate of the mask for each frame of the sequence; it encourages the mask in frame $p$ to align with the mask in frame $p+1$ after being warped by the optical flow from frame $p$ to frame $p+1$, and vice versa, while keeping the masks close to their initialization.

This approach can in fact be derived from a formulation of the problem in terms of spectral clustering, which also provides a way to compute initial estimates~$\maskini{p}$. We present this formulation and the derivation of our approach below.

\subsection{Vanilla Spectral Clustering for Segmentation}
\label{sec:sc}

We first briefly present spectral clustering, in the context of image segmentation as done in \cite{Shi00,meila}, for example. We then discuss how it can be extended to video segmentation. The reader interested in more details about spectral clustering can refer to the tutorial in \cite{tutorial_ulrike}.

For image segmentation with spectral clustering, one considers an affinity matrix we will denote $\calA$. Each row and each column of $\calA$ corresponds to an image location. The coefficient $\calA_{ij} \geq 0$ on row $i$ and column $j$ should express how likely image location for row $i$ and image location for column $j$ belong to the same cluster. In the following, we will identify the row and column indices $i$ and $j$ with the corresponding image locations for simplicity.

According to spectral clustering theory, a good segmentation mask can be obtained from the second largest eigenvector $\bXtwo$ of the normalized affinity matrix $\calW$ defined as 
\begin{equation}
    \calW = \calD^{-1} \calA \> ,
    \label{eq:WeqinvDA}
\end{equation}
where $\calD = \text{diag}( \{ \sum_{j} \calA_{ij} \}_i)$ is the degree matrix.
The coefficients of $\bXtwo$ correspond to image locations. 
By thresholding them,
one obtains a binary mask for the segment.

To extend this approach to object segmentation in videos, we also start from an affinity matrix $\calA$. Each row of $\calA$, and each column, corresponds to an image location in a video frame. We will denote the coefficients of $\calA$ with $\calA_{(pi)(qj)}$: $(pi)$ is a notation for the index of row corresponding to image location $i$ in frame $p$; similarly $(qj)$ is the index for column corresponding to image location $j$ in frame $q$.  Coefficients $\calA_{(pi)(qj)}$ should express how likely image location $i$ in frame $p$ and image location $j$ in frame $q$ are to correspond to the same object. For this, we will use the similarity between their local image features and optical flow.

However, this results in a very large $\calA$ matrix, and thus also a very large $\calW$ matrix. 
of size $NHW \times NHW$, which is in the order of $10^9 \, \times \, 10^9$ 
for typical sequences in standard benchmarks. This is clearly too large in practice both for storage and for eigenvector computation. Some methods exploit superpixels~\cite{fabio2012video} or graph sparsification~\cite{khoreva2015classifier} to decrease the computational complexity, but it complicates the approach; our method is more direct and more scalable.  Another approach~\cite{meila} is to retrieve the second largest eigenvector $\bXtwo$ with the classical problem:
\begin{equation} 
\bXtwo \,{=}\, \argmax_{\bX} \bX^\top \calW \bX \text{ \ such that \ } \|\bX\|_2 \,{=}\, 1 \> ,
\label{eq:X2argmaxXTWX} 
\end{equation}
which could be approximately but efficiently computed using Power Iteration Clustering~(PIC)~\cite{power-iteration}. However, it still cannot be scaled for common video \renaud{lengths}.

Nevertheless, as described below, we draw on this approach: Instead of building $\calW$ explicitly and struggling to compute its second largest eigenvector $\bXtwo$, we compute eigenvectors $\maskini{p}$ by PIC for each frame $p$ independently, for scalability. We use data that include inter-frame information to reinforce temporal consistency which leads to greater accuracy. \georgy{This makes our initialization scheme similar to TokenCut~\cite{wang2022tokencut}, except that we utilize approximate eigenvector extraction instead of full spectral decomposition as well as inclusion of optical flow connectivity between adjacent frames. In Table~\ref{tab:ablation-gains}(b), we show differences between using the baseline TokenCut method and our approximation without optical flow.} These initial masks are refined using additional inter-frame consistency constraints, resulting in an optimization problem that is much more tractable than the originating spectral clustering problem.

In the following, we introduce a suitable affinity matrix $\calA$, that we use in two ways: to efficiently compute good initial estimates $\maskini{p}$, and to derive the formulation in Eq.~\eqref{eq:obj} as
a simplification of term
$\bX^\top \calW \bX$ we wish 
to maximize.

\subsection{Affinity Matrix for Video Object Segmentation}
\label{sec:matrices}

Our affinity matrix $\calA$ relies both on object appearance features and on optical flow features:

\paragraph{Appearance features. } For these, we use an image feature extractor that generates an appearance feature vector $\dino{p}{i}$ of the object at location $i$ of frame $p$.  The idea is that the features at different locations of the same object should tend to look alike, while differing from the features located on other objects or on the background. Such appearance features can typically be learned using a dataset of annotated objects or, as in our case, using a self-supervised method such as DINO~\cite{caron2021emerging} or MoCo~\cite{chen2021mocov3}.

\paragraph{Optical flow features. } We use an image flow extractor that yields the optical flow $\flow{p}{q}{i} \,{\in}\, \IR^2$ at location $i$ between frame $p$ and frame $q$: What is seen at location $i$ in frame $p$ is also seen at location $j \,{=}\, i+\flow{p}{q}{i}$ in frame $q$. Such an optical flow can typically be obtained using various gradient-based formulations~\cite{Fleet2006opticalflow}, or learned using a dataset of annotated flows~\cite{teed-eccv20-raft,Sun2018PWC-Net,dosovitskiy-cvpr15-flownet,flownet2}. It can also be obtained, as in our case, using a self-supervised method such as ARFlow~\cite{liu2020learning}.

\paragraph{Affinity matrix. } We define our affinity matrix as follows: 
\begin{equation}
    \calA = 
    \left[
    \begin{array}{c@{\;}c@{\;}c@{\;}c@{\;}c}
         \aff{1}  & \rflw{1}    & 0       & \dots     & 0      \\
         \flw{1}  & \aff{2}    & \rflw{2} & \ddots    & \vdots \\
         0        & \flw{2}    & \aff{3} & \ddots    & 0      \\
         \vdots   & \ddots     & \ddots  & \ddots    & \rflw{N-1} \\
         0        & \dots      & 0       & \flw{N-1} & \aff{N} \\        
    \end{array}
    \right] \> .
    \label{eq:full_matrix}
\end{equation}
Each block is a $HW \times HW$ matrix. It contains the affinities between image locations in a frame and image locations in the same frame or in another one. We detail below matrices $\aff{p}$, $\flw{p}$ and $\rflw{p}$, and justify the use of $0$ matrices.

\paragraph{Matrices $\aff{p}$. } Each matrix $\aff{p}$ on the block diagonal of $\calA$ contains the affinities between two image locations in the same frame $p$. % Note that 
It is based not only on object appearance features within frame $p$, but also on information from the preceding and succeeding frames $p\,{-}\,1$ and $p\,{+}\,1$. Concretely, we take the affinity $\aff{p}^{i,j}$ between locations $i$ and $j$ in frame $p$ as the following combined similarity of their object appearance features and their forward and backward flows:
\begin{eqnarray}
\aff{p}^{i,j} &=&  \frac{1}{\alpha_\psi \,{+}\, 2\alpha_\phi} \,g_s\!\bigg(
\alpha_\psi \sdotprod{\dino{p}{i}}{\dino{p}{j}} + \nonumber\\
&&\!\!\!\!\!\!\!\!\!\!\!\!\!\!\!\alpha_\phi \left( \sdotprod{\flow{p}{p+1}{i}}{\flow{p}{p+1}{j}} + \sdotprod{\flow{p}{p-1}{i}}{\flow{p}{p-1}{j}} \right)\bigg)  \> ,
\label{eq:single-frame-affinity}
\label{eq:defA}
\end{eqnarray}
where $\alpha_\psi$ and $\alpha_\phi$ are relative weights, and
$g_s(\cdot)$ is a thresholding function that zeroes values lower than $s$. We use $s > 0$ to guarantee that the coefficients of $\aff{p}$ are positive.

\paragraph{Matrices $\flw{p}$ and $\rflw{p}$. } Each matrix $\flw{p}$ stores affinities between image locations in frame $p$ and image locations in frame $p\,{+}\,1$. Here, we only consider the optical flow between frames $p$ and $p\,{+}\,1$, and ignore the information given
appearance features. (Using appearance features as well might improve the results, but it would lead to a much more complex optimization problem.)
Two locations $i$ and $j$ in frames $p$ and $p\,{+}\,1$ are likely to both belong to the same object or both lie on the background if the optical flow maps one to the other. In that case, their affinity $\flw{p}^{i,j}$ should be large, close to 1; otherwise, we set it to 0. We thus take: 
\begin{equation}
\begin{array}{rcl}
\flw{p}^{i,j} & = & 
\left\{
\begin{array}{ll}
1 & \text{ if \ } i + \nint{\flow{p}{p+1}{i}} = j \\
0 & \text{ otherwise}
\end{array}
\right. \> ,
\end{array}\label{eq:Fdef}
\end{equation}
where $\nint{\flowsymb}$ is the vector of % the 
nearest integers to $\flowsymb$.
$\rflw{p}$ is defined likewise using the optical flow from frames $p\,{+}\,1$ to~$p$.

If $\mask{}$ denotes a 2D mask in vector form, then $\flw{p}\,\mask{}$ is ``\emph{almost}'' the mask $\warp{p}{p+1}(\mask{})$ after warping by the optical flow from frame $p$ to frame $p+1$, up to the integer discretization in Eq.\,\eqref{eq:Fdef}. It also applies to $\rflw{p}\,\mask{}$ and $\warp{p+1}{p}(\mask{})$.

The spectral clustering purist may notice that this expression for $\flw{p}$ and $\rflw{p}$ makes the affinity matrix $\calA$ not symmetric. However, $\rflw{p}$ is \emph{almost} equal to $\flw{p}^\top$; they are slightly different because (1)~optical flow is not exactly a bijection, as some pixels can appear or disappear, and (2)~the predicted flow from frame $p$ to $p+1$ is not exactly the inverse of the predicted flow from $p+1$ to $p$. In practice, we may consider that the affinity matrix $\calA$ is ``\emph{almost}'' symmetric.

\paragraph{Matrices $0$. }  Matrices that do not belong to the tri-diagonal of matrix $\calA$ contain affinities between image locations in two ``distant'', \emph{i.e.}, non-consecutive frames. Here, we ignore the information provided by the appearance features and the optical flow, resulting in the zero matrix. Being able to exploit the appearance features
would probably slightly improve the final segmentation, but would also make the optimization significantly \renaud{less scalable and} more complex. Ignoring the optical flow \renaud{between distant frames} is a safe thing to do \renaud{anyway, as its estimation is likely} 
unreliable.

\subsection{Deriving the Objective Function}
\label{sec:derivations}

By expanding the term $\bX^\top\calW\bX$ found in Eq.\,\eqref{eq:X2argmaxXTWX}, using the definition of the normalized affinity matrix $\calW$ in Eq.\,\eqref{eq:WeqinvDA}, the notation $\bX$ for its second eigenvector, and the definition of the affinity matrix $\calA$ in Eq.\,\eqref{eq:full_matrix}, we obtain: 
\begin{eqnarray}
    \bX^\top\calW\bX &=& 
\sum_p \mask{p}^\top D_p^\inv\aff{p}\mask{p} \; + \nonumber\\
&&\!\!\!\!\!\!\!\!\!\!\!\!\!\!\!\!\!\!\!\!\!\!\!\!\!\!\!\!\sum_p \mask{p+1}^\top D_{p+1}^\inv \flw{p}\mask{p} \; + 
\sum_p \mask{p}^\top D_p^\inv\rflw{p}\mask{p+1} \> ,
    \label{eq:XWX}
\end{eqnarray}
where the $D_p$ matrices are on the block diagonal of~$\calD$ and are themselves diagonal.

Computing this expression
seemingly involvesx
the products between matrices $\aff{p}$, $\flw{p}$ and vectors $\mask{p}$. It would require large amounts of memory and be prohibitively slow as these matrices are very large. However, we show that we do not need to store these matrices, nor do we need to compute these products for estimating masks $\mask{p}$.

The terms $\mask{p}^\top D_p^\inv\aff{p}\mask{p}$, when considered independently, lead to a spectral clustering problem per frame. The second eigenvector $\maskini{p}$ of matrix $W_p = D_p^\inv\aff{p}$ is in practice a good first estimate of $\mask{p}$, thanks to the combination of the \renaud{object} appearance features and the optical flow. 

In the supplementary material, we also show that when $\mask{p}$ is close to $\maskini{p}$, then:
\begin{equation}
\mask{p}^\top D_p^\inv \aff{p} \mask{p} \approx \left({\maskini{p}}\right)^\top \mask{p} \> .
\label{eq:xzero}
\end{equation}

Furthermore, as mentioned in Section~\ref{sec:matrices}, the terms $\flw{p}\mask{}$ and $\rflw{p} \mask{}$ are approximations of the warpings of vector $\mask{}$ by the optical flow, \renaud{i.e., $\flw{p}\mask{} \approx \warp{p}{p+1}(\mask{})$ and $\rflw{p}\mask{} \approx \warp{p+1}{p}(\mask{})$.  As matrix $D_p$ is diagonal}, we can compute efficiently the last two sums in Eq.~\eqref{eq:XWX} using
\begin{equation}
\begin{array}{l}
  \mask{p+1}^\top D_{p+1}^\inv \flw{p}\mask{p} = \mask{p+1}^\top  \left(\bd_{p+1} \odot \warp{p}{p+1}(\mask{p})\right) \>, \\[0.5em]
  \mask{p}^\top D_p^\inv \flw{p}^\top \mask{p+1} = \mask{p}^\top \left(\bd_p \odot \warp{p+1}{p}(\mask{p+1})\right) \>, \\
\end{array}\label{eq:warp}
\end{equation}
where $\bd_p$ denotes the \renaud{vectorized} coefficients on the diagonal of matrix $D_p^\inv$ and $\odot$ is the element-wise product. The terms $\bd_p$ weight the warped masks. We noticed during our experiments that their influence was very limited, and we did not keep them in our objective function for simplicity.

Moreover, while spectral clustering is a powerful framework, it actually is an approximation as it relaxes the search for binary masks by looking for continuous values for the coefficients of $\bX$, instead of binary ones. To encourage the optimization to recover binary values, we use the cross-entropy \renaud{(minimizing it)} when measuring the similarity between masks, rather than the dot product \renaud{(maximizing it)}.

\vincent{
From Eqs.\,\eqref{eq:X2argmaxXTWX}, \eqref{eq:XWX}, \eqref{eq:xzero}, \eqref{eq:warp}, we finally obtain the formulation in Eq.\,\eqref{eq:obj}.}
In practice, to minimize $\calL(\{\mask{p}\}_p)$, and thus maximize the flow consistency while not deviating too much from initial estimates, we first compute the eigenvectors $\maskini{p}$ and use them to initialize the vectors $\mask{p}$. Note that since the $\mask{p}$ vectors remain close to the $\maskini{p}$ vectors, the norm of $\bX$ remain approximately constant during optimization and the unit constraint \renaud{in Eq.~\eqref{eq:X2argmaxXTWX}} is approximately satisfied. \renaud{After convergence of the minimization, the soft masks $\mask{p}$ are discretized using $K$-means with $K\,{=}\,2$ to separate the object of interest from the background.} Implementation details are provided in the supplementary material.

\section{Experiments}
\label{sec:experiments}
In this section, we first introduce the implementation details of our method. Then we describe the datasets we use for comparison with other methods and make a comprehensive analysis of each component in our approach.

\subsection{Datasets}

We evaluate our model on three standard benchmarks in unsupervised video object segmentation: \relax{DAVIS2016}, \relax{SegTrack-v2}, and \relax{FBMS59}. \textbf{DAVIS2016}~\cite{perazzi-cvpr16-abenchmarkdataset} is a densely-annotated video object segmentation dataset, featuring 50 sequences and 3455 frames in total, captured in 1080p resolution at 24 FPS with precise annotation of a primary moving object at 480p. \textbf{SegTrack-v2}~\cite{li-iccv13-videosegmentation} is a densely-annotated dataset of 14 sequences with 976 frames in total. It sometimes features multiple objects in a single video and has multiple challenges such as motion blur, deformations, interactions, and objects being static. \textbf{FBMS59}~\cite{Brox14,brox2010object} is a dataset of 59 sequences with every 20-th frame being annotated, which yields 720 annotated frames in total. The dataset may involve multiple objects (some of which can be static), occlusions, and other challenging conditions. Since  SegTrack-v2 and FBMS might feature multiple objects in one scene and since our method is interested in the main object segmentation, we merge the segmentation masks of the individual objects into one, similar to \cite{fusionseg,yang2019unsupervised,yang2021self}.

\subsection{Metrics}

\textbf{Jaccard ($\calJ$).} For all datasets, we evaluate them with the Jaccard metric $\calJ$, which measures the intersection-over-union between predicted masks and ground-truth masks. 

\textbf{Contour accuracy ($\calF$).} For the DAVIS2016 dataset, we also report contour accuracy. This measure treats the mask as a set of closed contours to calculate their precision and recall with respect to the annotation. The contour accuracy is then taken as $\calF = \frac{2 P_c R_c}{P_c + R_c}$ where $P_c$ is the contour precision and $R_c$ is the contour recall.  

\textbf{Computational cost.} On an Nvidia V100, our initialization and optimization runs \renaud{on average} in $\sim$\,0.5\,s/frame, where initial eigenvector extraction takes $\sim$\,0.1\,s/frame and the optimization takes $\sim$\,0.4\,s/frame.

Optimization consumes $\sim$\,8\,GB of VRAM for typical sequences in DAVIS2016, and $\sim$\,20\,MFLOPS/frame. 

\newcommand{\Opt}{Optimization (Eq.~\eqref{eq:obj})}

\setlength\textfloatsep{4mm} 
\begin{table}[t]
\centering
\begin{tabular}{c}
\scalebox{0.68}{\hspace*{-3mm}%
    \begin{tabular}[b]{lc@{~}rc@{~}rc@{~}r@{}}
    \toprule 
    \renaud{{\bf (a) Different appearance features}} & \multicolumn{2}{c}{($\calJ \uparrow$)}  & \multicolumn{2}{c}{($\calF \uparrow$)}   \\
    \midrule
    (BT [RN-50] + ARFlow) & 53.5 && 30.3 \\
    (BT [RN-50] + ARFlow) + \Opt 
    & 59.5 &(\textbf{+6.0}) & 48.2  &(\textbf{+17.9}) \\
    \midrule
    (SWAV [RN-50] + ARFlow) & 48.0 && 27.2\\
    (SWAV [RN-50] + ARFlow) + \Opt 
    & 53.6 &(\textbf{+5.6}) & 46.0  &(\textbf{+18.8})\\
    \midrule
    (MoCo-v3 [ViT] + ARFlow) & 58.0 && 35.3 \\
     (MoCo-v3 [ViT] + ARFlow) + \Opt       & 64.2 &(\textbf{+6.2}) & 61.1 &(\textbf{+25.8}) \\
    \midrule
    (DINO [ViT] + ARFlow)       & 72.1 && 72.5 \\
     (DINO [ViT] + ARFlow) + \Opt       & 76.8 &(\textbf{+4.7}) & 77.0 &(\textbf{+4.5}) \\
    \toprule 
    \renaud{{\bf (b) Different optical flows}} & \multicolumn{2}{c}{($\calJ \uparrow$)}  & \multicolumn{2}{c}{($\calF \uparrow$)}   \\
    \midrule
     DINO  & 61.2 && 65.8  \\
     TokenCut~\cite{wang2022tokencut}  & 62.7 && 62.3 \\
     DINO + \Opt & 66.7 &(\textbf{+5.5}) & 70.4 &(\textbf{+4.6})\\
    \midrule
   (DINO + RAFT)         & 70.7 && 72.9 \\
    (DINO + RAFT) + \Opt         & 75.3 &(\textbf{+4.6}) & 76.2  &(\textbf{+3.3})\\
    \midrule
     (DINO + ARFlow)       & 72.1 && 72.5 \\
    (DINO + ARFlow) + \Opt & 76.8 &(\textbf{+4.7}) & 77.0 &(\textbf{+4.5}) \\
    \bottomrule
  \end{tabular}
  }
\end{tabular}
\vspace*{-3mm}
\caption{\label{tab:ablation-gains}
\nermin{\textbf{Ablation experiments on DAVIS2016 for different optical flow and appearance features\vincent{ for the frame-based initialization and after optimization}.} (a) Effects of different self-supervised methods for appearance features. (b) Effects of optical flow and different optical flow methods on performance. We obtain all initial results using Spectral Clustering on each frame independently. “+Optimization” denotes the results after optimizing our objective function. No post processing is applied to the results. Our optimization approach consistently improves the segmentation results by a large margin for all appearance features, with or without optical flow for initializing the masks. 
}}
\end{table}

\subsection{Ablation and Parameter Study}

In order to understand the different factors that contribute to the performance of our method, we conduct a series of ablation studies. % discussed in this section.
Table~\ref{tab:ablation-gains} reports the importance of the different components of our pipeline. Note that we do not apply any post-processing in our ablation experiments to show direct gains by our method. We evaluate the different aspects of our method as detailed below.

\textbf{Initial masks $\maskini{p}$ from appearance only.} 
We study applying spectral clustering to appearance features only.
We consider recent self-supervised features: DINO~\cite{caron2021emerging}, MoCo-v3~\cite{chen2021mocov3}, SWAV~\cite{caron2020unsupervised} and Barlow Twins~(BT)~\cite{zbontar2021barlow}. \georgy{Table~\ref{tab:ablation-gains}(a) compares how these appearance features affect our results. 
\nermin{We obtain the best performance with DINO. The fact that DINO largely outperforms other self-supervised features, as also noted in~\cite{wang2022tokencut, melaskyriazi2022deep}, remains to be understood, but is out of scope. We also note that our method can exploit any image  features. Better feature extractors in the future could even improve our results further.} \nermin{Table~\ref{tab:ablation-gains}(a) also shows that,} in every case our optimization method improves the $\calJ$ metric by the average value of \textbf{+5.6\%} and the $\calF$ metric by the average value of \textbf{+16.6\%}. This ablation also shows one potential application of our method, which could serve as a way to benchmark fine-grained representation capacity of self-supervised features~\cite{wang2021different,Ericsson2021HowTransfer}: Given frozen features and strictly predefined optical flow, one can "plug-in" new features and estimate how good the representation capacity is.}

As it has the most similar (graph-based) formulation to ours, we also compare to TokenCut~\cite{wang2022tokencut} applied to each frame independently, using the same DINO features. On David2016, our mask extraction from single frames (i.e., our initial masks $\maskini{p}$) is 1.5 pt behind TokenCut on the $\calJ$ metric, while outperforming it on~$\calF$ by 3.5 pts, which shows that our initialization is better at detecting object boundaries on individual images. More importantly, we are $\sim 170\times$ faster than TokenCut~\cite{wang2022tokencut} (0.1\,s/frame vs.\ 17\,s/frame).

\textbf{Initial masks $\maskini{p}$ from both appearance and flow.} 
We apply spectral clustering on  the combination of appearance features and optical flow, as in Eq.\,\eqref{eq:single-frame-affinity}.
We consider supervised RAFT and self-supervised ARFlow. Effects of adding the optical flow to our approach can be seen in Table~\ref{tab:ablation-gains}(b). The addition of optical flow to the appearance features greatly improves single-frame clustering performance and gives us a good initialization for the object masks. (Two successive frames are used for computing the optical flow.)

\textbf{Masks $\mask{p}$ optimized from different initial masks $\maskini{p}$. } 
We present the effects of our global optimization on different appearance features and optical flows in Table~\ref{tab:ablation-gains}. Our optimization gives an average boost of \textbf{+4.9\%} over the single frame clustering, validating our approach in all cases.

\textbf{Number of flow steps.} We found (see full results in supp.~mat.) that one flow step achieves the best performance, which supports our assumption of using a tridiagonal global affinity matrix: Even in the single flow step case, all frames are tied together via optical flow. Conversely, more steps might complicate the optimization due to the potentially noisier  flow estimated between distant frames because of the larger displacements and additional occlusions.

\textbf{Cross-entropy vs.\ dot product.} In our objective function, we use the cross-entropy rather than the dot product between the masks. We compared the two options experimentally and found that using the cross-entropy indeed improves the masks that we recover. We provide the full results in the supplementary material.

\begin{table}[t]
  \centering
  \resizebox{1\columnwidth}{!}{
    \begin{tabular}{@{}lccc|cccc@{}}
    \toprule
    & \hspace{-6mm}Training\hspace{-4mm} & \hspace{-6mm}Optical\hspace{-2mm} & \hspace{-6mm}Fully & \multicolumn{2}{c}{DAVIS} & STv2 & FBMS59 \\
    Method & \hspace{-6mm}on videos\hspace{-4mm} & \hspace{-2mm}flow & \hspace{-8mm}unsuperv.\hspace{-1mm} & $\calJ \uparrow$ & $\calF \uparrow$ & $\calJ \uparrow$ & $\calJ \uparrow$\\
    \midrule
    \multicolumn{4}{l}{$\!\!\!\!\!$\textbf{\textit{(Mostly) Unsupervised methods:}}} & & & &\\ 
 CUT~\cite{keuper2015motion}  &    & LDOF~\cite{brox2010large} &\cmark& 55.2 & 55.2 & 54.3 & 57.2 \\
FTS~\cite{papazoglou2013fast}  &   &  LDOF~\cite{sundaram2010dense}  &\cmark& 55.8 & 51.1 & 47.8 & 47.7 \\
AMD~\cite{liu2021emergence}  &   \cmark  &  
 &\cmark& 57.8 & - & 57.0 & 47.5 \\
   MoSeg~\cite{yang2021self}  &   \cmark  &   RAFT~\cite{teed-eccv20-raft}  && 68.3 & 61.1 & 58.6 & 53.1 \\
CIS~\cite{yang2019unsupervised}    &  \cmark &  PWCNet~\cite{Sun2018PWC-Net} && 71.5 & 70.5 & 62.0 & 63.5 \\
DS~\cite{Ye2022CVPR}    &  \cmark &  RAFT~\cite{teed-eccv20-raft} && 79.1 & - & 72.1 & 71.8 \\
DyStaB~\cite{yang2021dystab} & \cmark & PWCNet~\cite{Sun2018PWC-Net}  && 80.0 &  -  &  74.2 &  \textbf{73.2} \\
    \textbf{Ours}  &  &  ARFlow~\cite{liu2020learning} &\cmark& \textbf{80.2} & \textbf{77.5} & \textbf{74.9} & 70.0 \\
    \midrule
    \multicolumn{4}{l}{$\!\!\!\!\!$\textbf{\textit{Supervised methods:}}} & & & &\\ 
        NLC~\cite{Faktor2014VideoSB}   &    &  SIFTFlow~\cite{siftflow} &&  55.1 & 52.3 & 67.2 & 51.5 \\
            SFL~\cite{cheng2017segflow}  & \cmark & FlowNetS~\cite{flownet}  && 67.4 & 66.7 & - & -  \\
    FSEG~\cite{fusionseg}  & \cmark &   && 70.7 & 65.3 & 61.4 & 68.4  \\
    LVO~\cite{tokmakov2017learning}  & \cmark & MP-Net~\cite{mpnet} && 75.9 & 72.1 & 57.3 & 65.1  \\
           ARP~\cite{arp}   &  & CPMFlow~\cite{cpmflow}  && 76.2 & 70.6 & 57.2 & 59.8  \\
   
    MSgStP~\cite{hu2018unsupervised} &  &  && 77.6  & - & 70.1 & 60.8  \\
    MATNet~\cite{zhou2020motion}   & \cmark  & PWCNet~\cite{Sun2018PWC-Net} && 82.4 & 80.7 & - & -  \\
  DyStaB~\cite{yang2021dystab} & \cmark & PWCNet~\cite{Sun2018PWC-Net} && 82.8 &  -  &  \textbf{74.2} & \textbf{75.8} \\
    3DC-Seg~\cite{mahadevan2020making}   & \cmark &  && 84.3 & 84.7  & - & - \\
    ViTAE~\cite{vitae2022} & \cmark & && 89.2 & 90.4 & - & - \\
    BATMAN~\cite{batman2022} & \cmark & RAFT~\cite{teed-eccv20-raft} && \textbf{90.7} & \textbf{94.2} & - & - \\
 \bottomrule
  \end{tabular}
  }
 \vspace*{-2mm}
  \caption{\label{tab:results-full}{\bf Results of our approach.} We show the performance of our approach on three standard benchmarks for video object segmentation (DAVIS2016, SegTrack-v2, and FBMS59), where our approach achieves state-of-the-art results despite. We also provide a comparison with some supervised methods, where we achieve a performance comparable to some of them without using any supervision. Note that recent unsupervised methods~(MoSeg, CIS, DyStaB and IKE) use supervised optical flow estimators such as RAFT~\cite{teed-eccv20-raft}, PWCNet~\cite{Sun2018PWC-Net} or FlowNet~\cite{flownet,flownet2}.    }
\end{table}

\subsection{Comparison to the State of the Art}
\label{sec:sota}

\nermin{Here, we use our best configuration~(DINO [ViT] + ARFlow  + Optimization). Following CIS~\cite{yang2019unsupervised}  and DyStaB~\cite{yang2021dystab}, we use a CRF as a post-processing step. Table~\ref{tab:results-full} presents the performances of our method and several established state-of-the-art unsupervised and supervised methods. Overall, our method performs on-par with the state-of-the-art methods on DAVIS2016 and it achieves the best performance on the STv2 dataset. Our method also has high contour accuracy $\calF$, which shows that our approach achieves high quality boundaries. Among unsupervised methods, our method outperforms all previous methods by achieving 80.2 $\calJ$ and 74.9 $\calJ$ scores on DAVIS2016 and STv2, respectively. Our method outperforms the SOTA method DyStaB albeit slightly, with a much simpler approach. Also note  that,  in  contrast  to  recent  unsupervised methods, our approach does not use any supervised component in the pipeline, including optical flow.} 

\nermin{Table~\ref{tab:results-full} shows the generalization ability of our method: Rather than training on a target dataset, we leverage general images features obtained from a network pretrained on a large dataset with no supervision. While it may not be optimal in some contexts due to different data distributions, we achieve an excellent performance on three different benchmarks, without training and reusing the exact same networks for feature extraction and flow computation.
}

Comparing ViT and CNN based methods is not straightforward as each of them has their own advantages. Although other methods do not exploit ViTs, the most recent ones do use advanced CNN networks, \eg, DeepLabv3 (used in DyStaB), one of the SOTA architectures for segmentation. Besides, while we rely only on pretrained networks, although possibly on large datasets (\eg, ImageNet for appearance, Sintel for flow), a number of other methods (supervised or unsupervised) are advantaged by the fact they train on the target datasets, hence on the task itself and using a data distribution closer to the test sets. Note that other methods in Table~\ref{tab:results-full} also use extra data beyond the evaluated dataset, \eg, DyStaB uses ImageNet-pretrained weights to initialize its network, while AMD uses Youtube-VOS (a large video dataset) pretrained weights.

\begin{figure}
  \centering
   \resizebox{1.0\columnwidth}{!}{%
  \begin{tabular}{@{}c@{\;}c@{\;}c@{\;}c@{}}
     \includegraphics[width=0.24\linewidth]{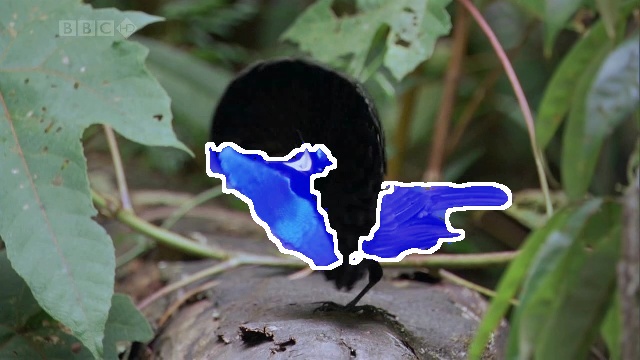} & \includegraphics[width=0.24\linewidth]{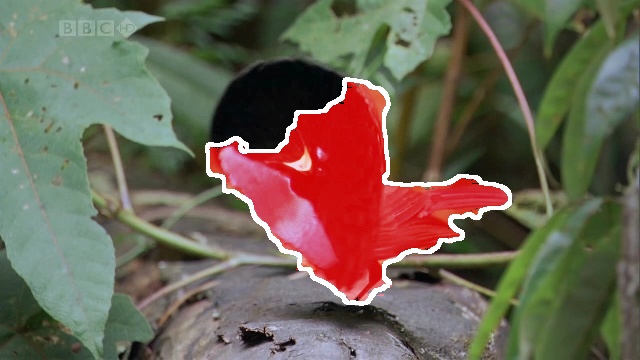} & \includegraphics[width=0.24\linewidth]{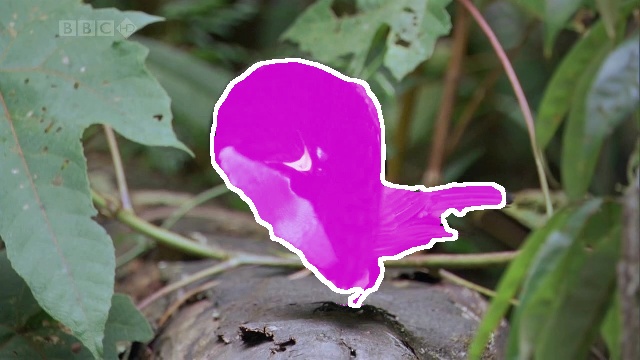} & \includegraphics[width=0.24\linewidth]{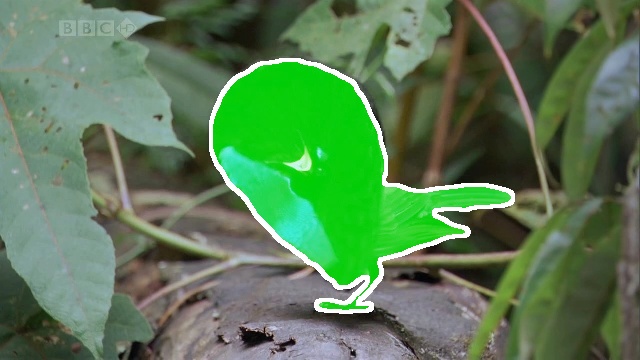} \\
 
     \includegraphics[width=0.24\linewidth]{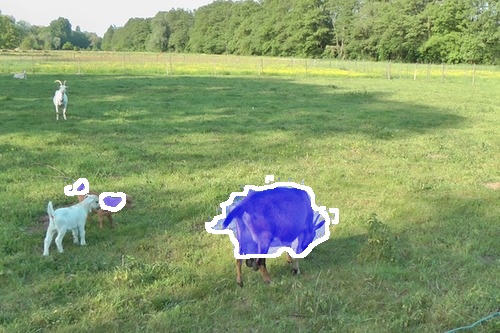} & \includegraphics[width=0.24\linewidth]{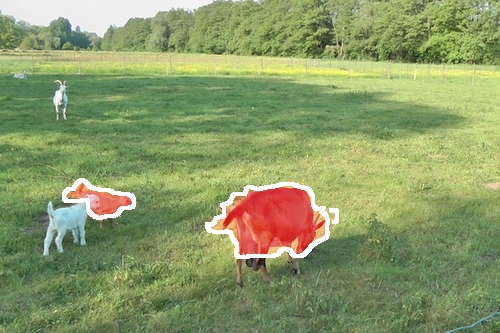} & \includegraphics[width=0.24\linewidth]{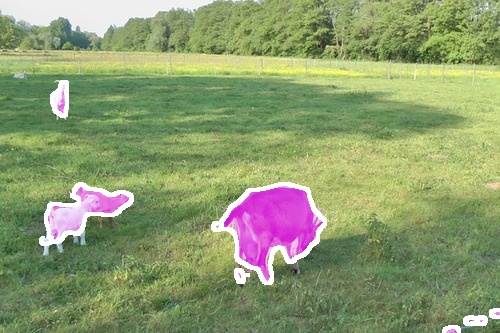} & \includegraphics[width=0.24\linewidth]{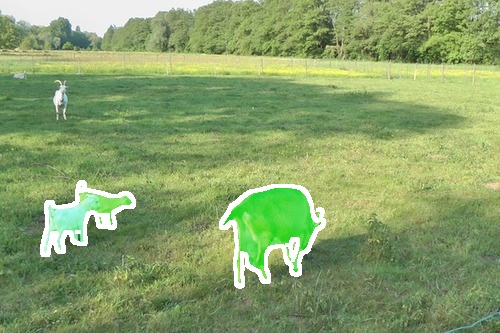} \\
    
    \includegraphics[width=0.24\linewidth]{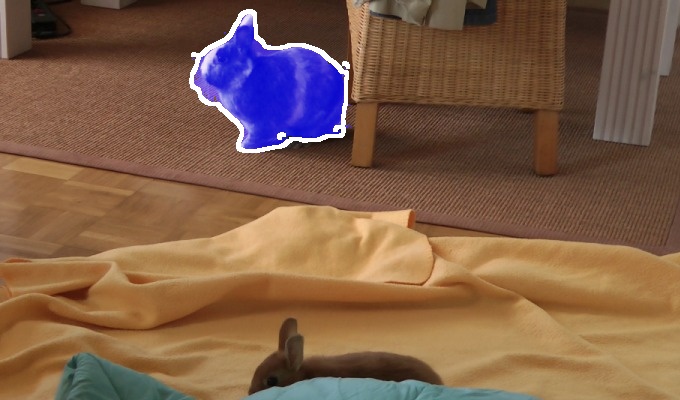} & \includegraphics[width=0.24\linewidth]{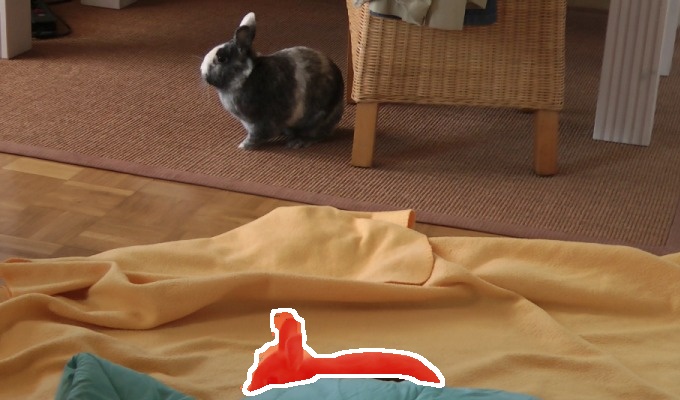} & \includegraphics[width=0.24\linewidth]{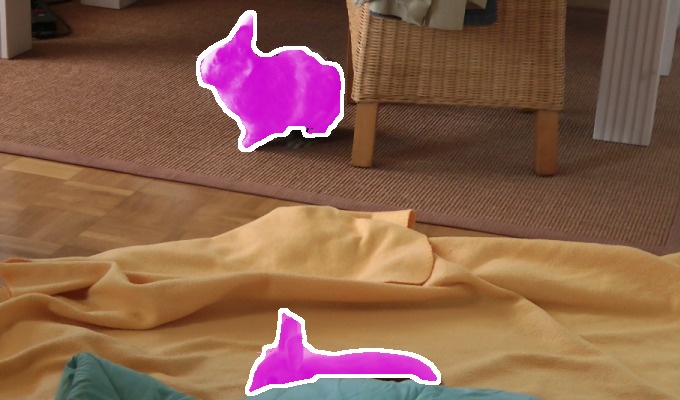} & \includegraphics[width=0.24\linewidth]{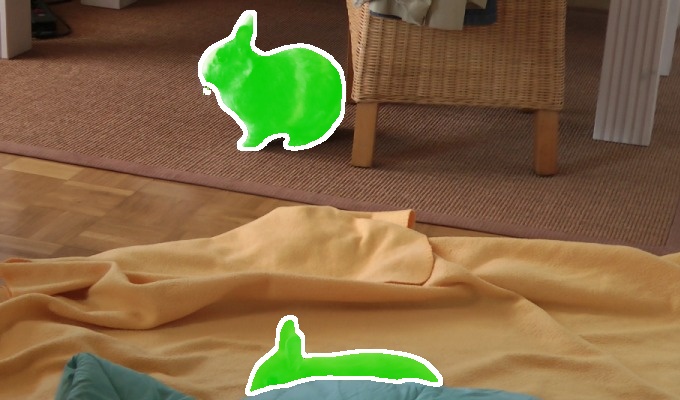} \\
    
    \includegraphics[width=0.24\linewidth,trim={8cm 4cm 5cm 3cm},clip]{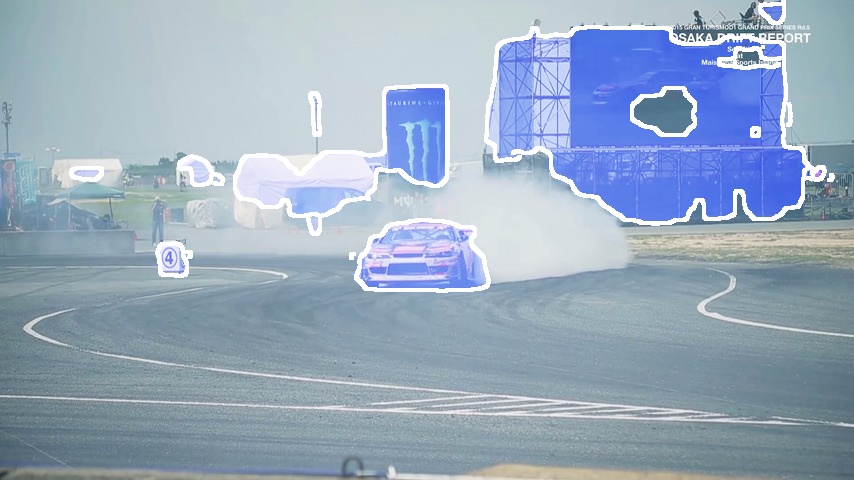} & \includegraphics[width=0.24\linewidth,trim={8cm 4cm 5cm 3cm},clip]{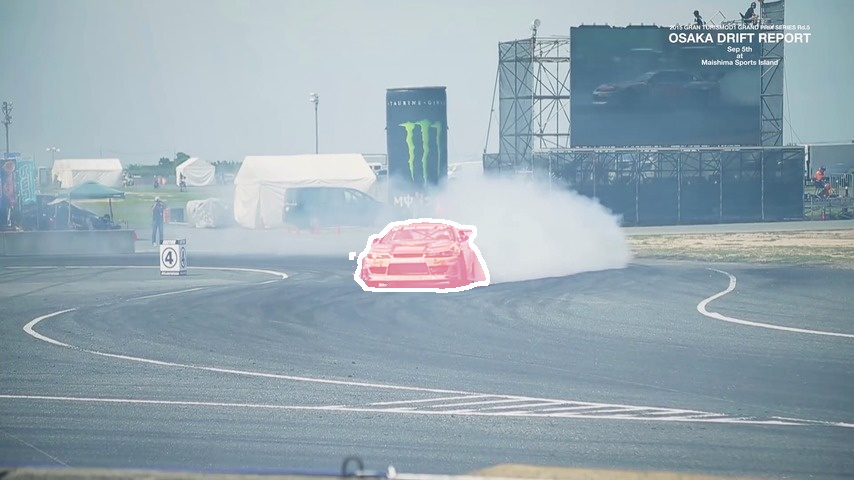} & \includegraphics[width=0.24\linewidth,trim={8cm 4cm 5cm 3cm},clip]{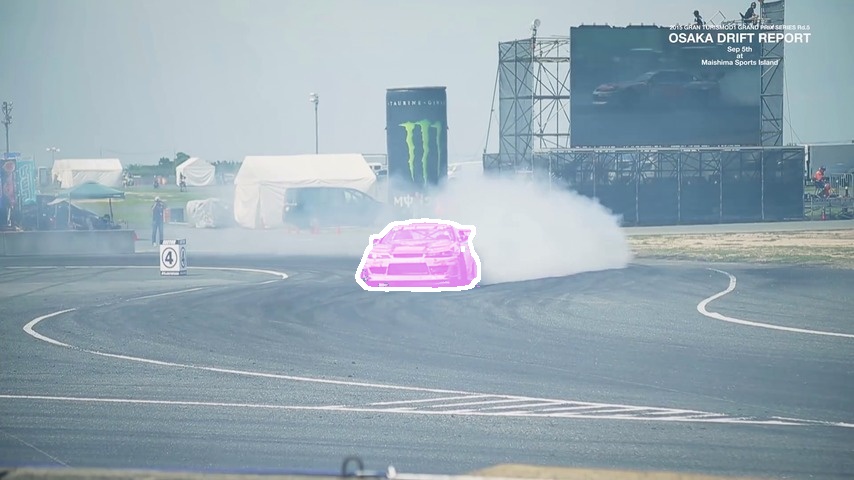} & \includegraphics[width=0.24\linewidth,trim={8cm 4cm 5cm 3cm},clip]{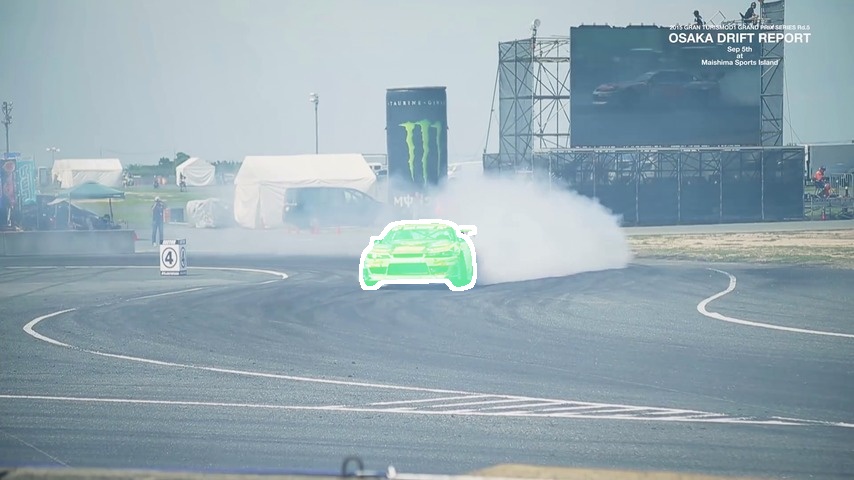} \\
    
    \includegraphics[width=0.24\linewidth,trim={3cm 1cm 3cm 1cm},clip]{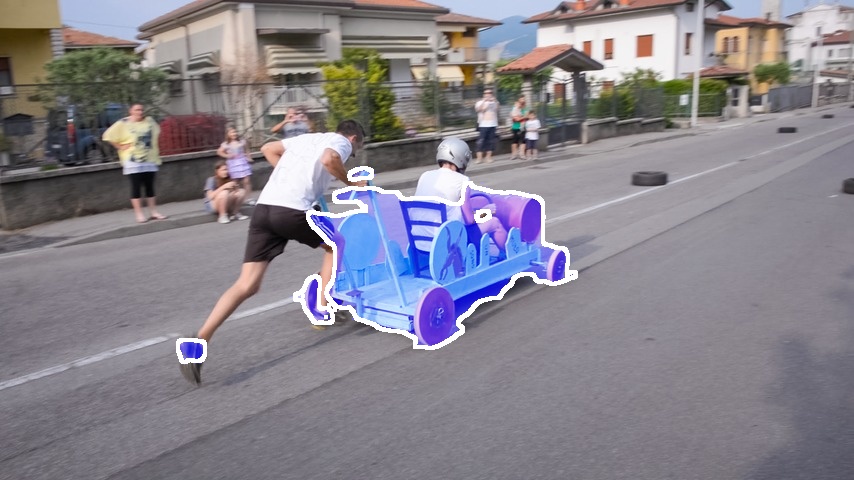} & \includegraphics[width=0.24\linewidth,trim={3cm 1cm 3cm 1cm},clip]{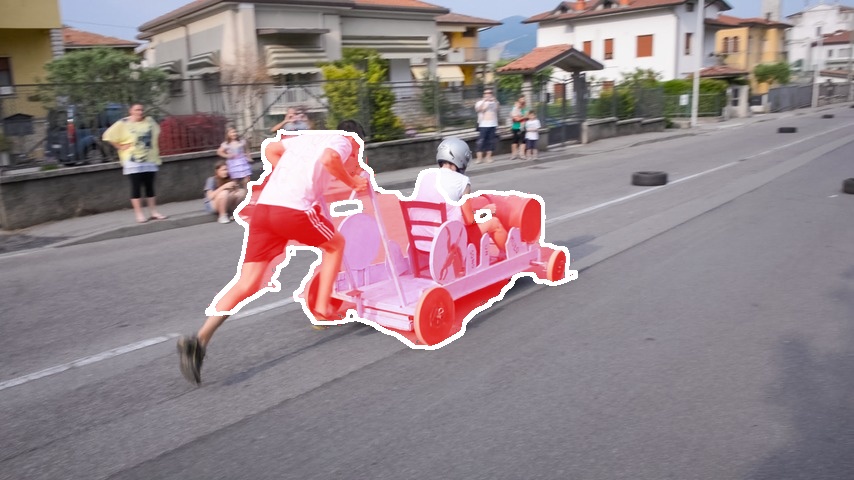} & \includegraphics[width=0.24\linewidth,trim={3cm 1cm 3cm 1cm},clip]{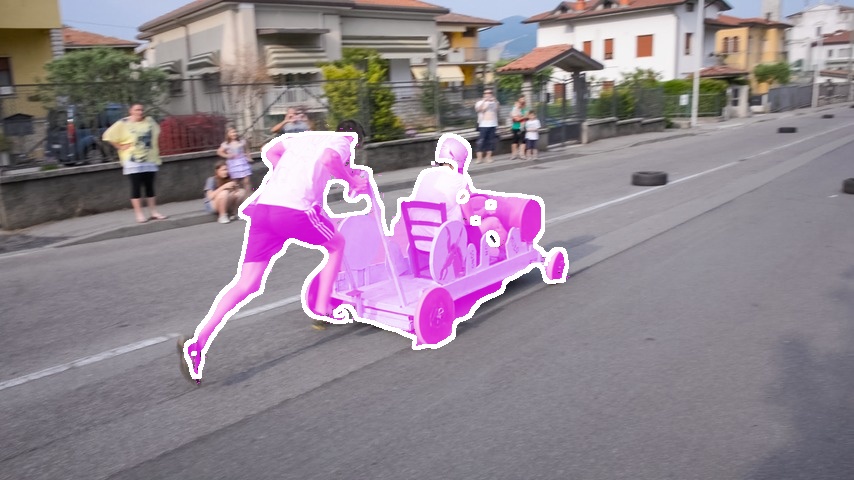} & \includegraphics[width=0.24\linewidth,trim={3cm 1cm 3cm 1cm},clip]{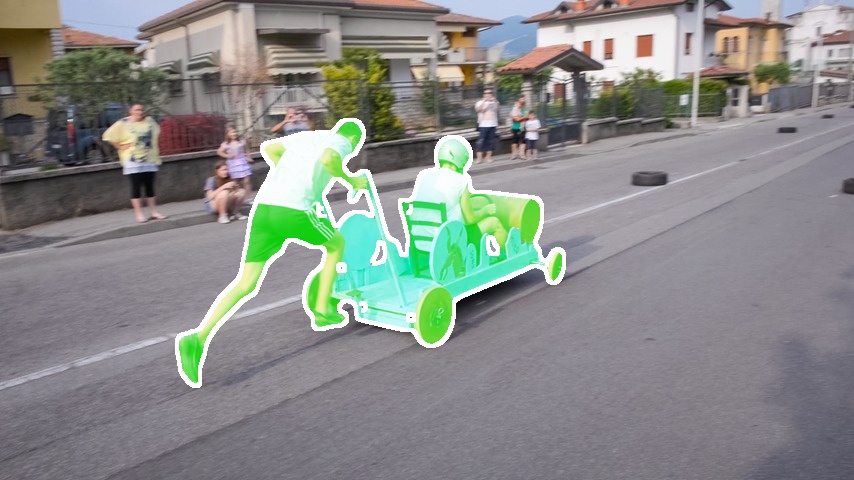} \\
    
    \includegraphics[width=0.24\linewidth]{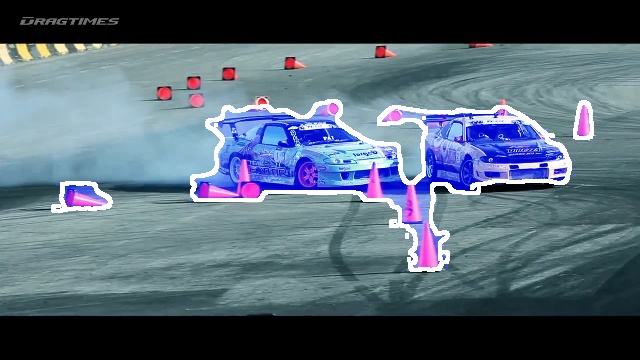} & \includegraphics[width=0.24\linewidth]{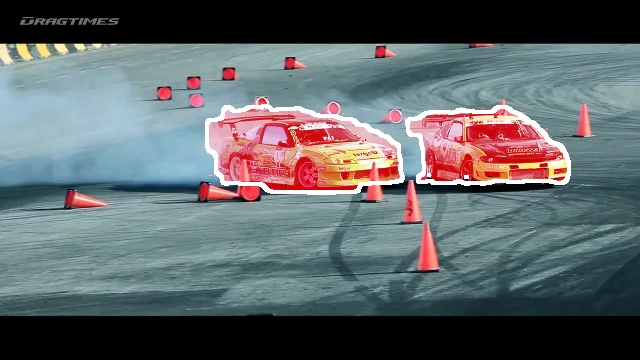} & \includegraphics[width=0.24\linewidth]{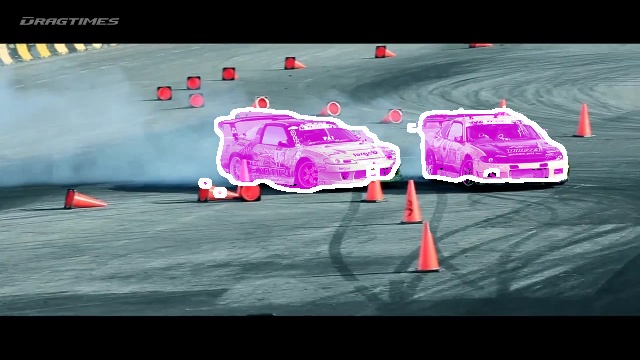} & \includegraphics[width=0.24\linewidth]{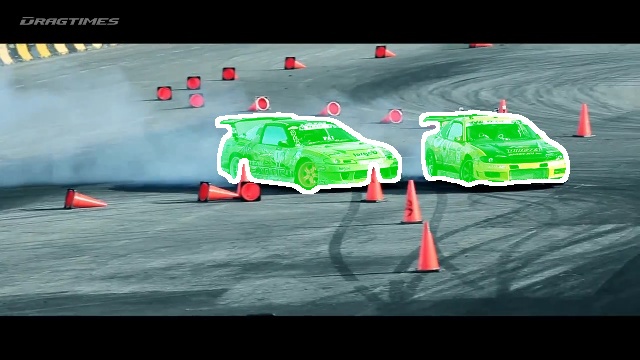} \\
    
    {\footnotesize DINO} & \footnotesize DINO + & \footnotesize DINO + & \footnotesize Ground Truth\\[-3pt]    
    & \footnotesize ARFlow & \footnotesize ARFlow + Opt & \\    
    
    % {\scriptsize DINO} & \scriptsize DINO + ARFlow & \scriptsize DINO + ARFlow + Opt & \scriptsize Ground Truth\\[-3pt]    

  \end{tabular}}
  \vspace*{-3mm}
  \caption{\label{fig:qualitative}{\bf Qualitative results} obtained with our approach. The quality of segmentation improves with each component. In the first row, partially segmented object is recovered; the second and third rows show that we successfully recover the small and occluded objects; in the fourth row our method removes the background residuals and finally in the fifth and sixth rows, we show that our method recovers and segments multiple close objects.  
  }
\end{figure}

Figure~\ref{fig:failures} shows some failure cases of our approach. More examples are provided in the supplementary material. Note that some of those failure cases can be removed by further postprocessing, but we do not use it to keep our approach simple. We show examples of segmentation results by our method for qualitative visual inspection in Figure~\ref{fig:qualitative}.

\begin{figure}
\vspace{-2mm}
  \centering
  \hspace{-3mm}\begin{tabular}{c@{\;}cc}
   \includegraphics[align=c,width=0.318\linewidth]{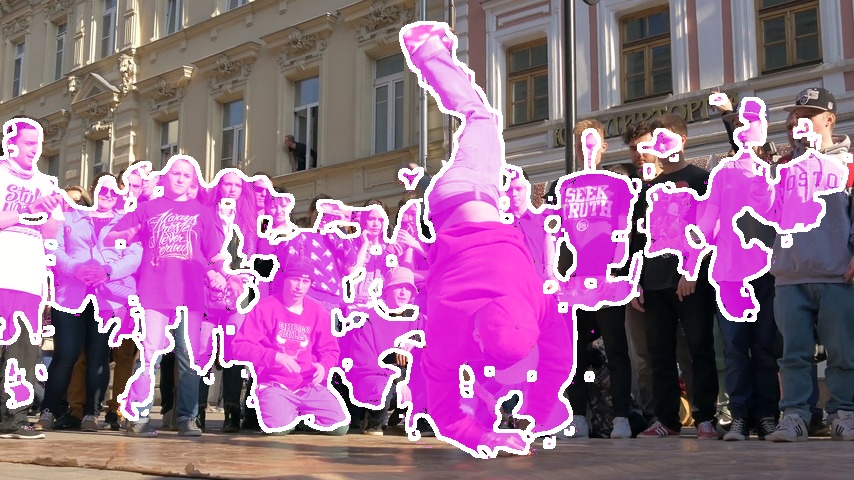} & \includegraphics[align=c,width=0.318\linewidth]{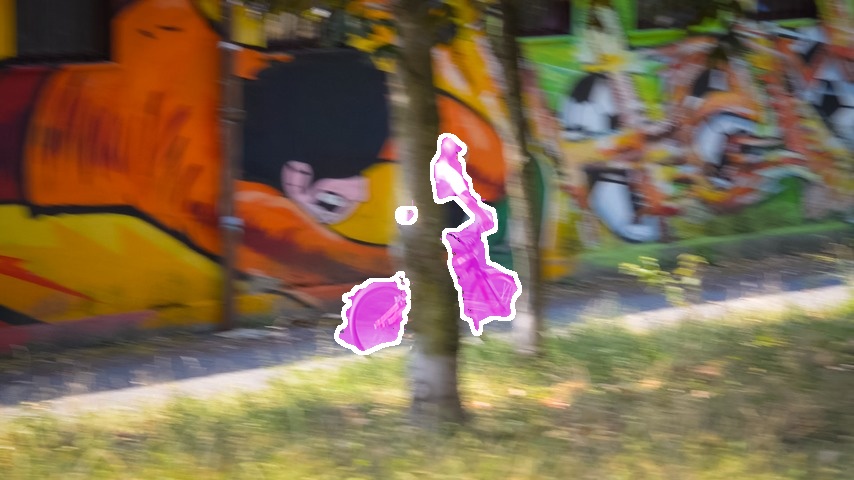} &
   \hspace*{-3.1mm}\includegraphics[align=c,width=0.318\linewidth]{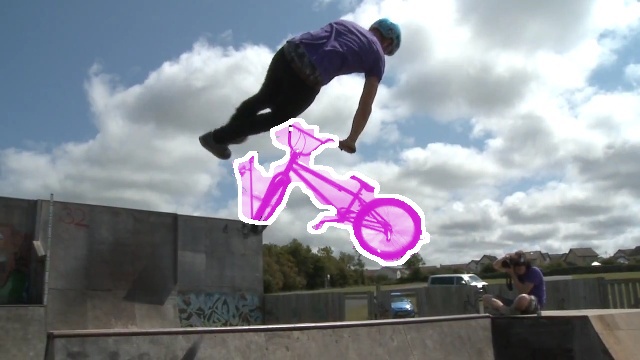}
  \end{tabular}
  \vspace{-1mm}
 \caption{\label{fig:failures}{\bf Failure cases.}  Our approach has 3 main failure modes: 
 \textit{Oversegmentation} which is most often caused by multiple similar objects in the scene, imprecise masks due to \textit{occlusion}, and \textit{undersegmentation} in scenes with multiple objects which leads to the parts of the primary object not being segmented. In the last example, note that when we use DINO features, our clusters tend to group a single semantic class, which is the desired behavior for object discovery, rather than segment anything that moves. This is thus a failure case w.r.t. the benchmark, but not to the goal of object discovery.}
\end{figure}

\section{Conclusion}
\label{sec:conclusion}
Our method consists of minimizing an objective function that is intuitive, simple to implement, and can be optimized efficiently. It can be derived from spectral clustering, which gives it solid theoretical ground. It could also be used to evaluate fine-grained capabilities of modern self-supervised representations, which is still a very active area of research~\cite{wang2021different,Ericsson2021HowTransfer}. We hope that 
the simplicity of our method and its connection to spectral clustering will provide others with insights for future development.

% We presented a simple method for object discovery and segmentation in a single videos, which relies only on pre-trained self-supervised methods for the extraction of visual features. Our method can be derived from spectral clustering, and consists of minimizing an objective function that is intuitive and simple to implement. It can also serve as an additional benchmark to evaluate fine-grained capabilities of modern self-supervised representations, which is still a very active area of research~\cite{wang2021different,Ericsson2021HowTransfer}. We hope that 
% the simplicity of our method and its connection to spectral clustering will provide others with insights for future development.

\section*{Acknowledgments}
This work was granted access to the HPC resources of IDRIS under the allocation 2021-AD011012896 made by GENCI and supported in part by the Chistera IPalm project.

{\small
\bibliographystyle{ieee_fullname}
\bibliography{wacv}
}

\newpage
\clearpage
\appendix

\section{Supplementary material -- Overview}

In this supplementary material:
\begin{itemize}
    \item we justify design choices and assumptions made in the main paper~(Sections~\ref{sec:usetridiagaff}-%
    \ref{sec:alleviating}); 
\item we show implementation details of our approach~(Section~~\ref{sec:implementation});
\item we provide additional failure cases~(Section~~\ref{sec:failcases});
\item we provide more qualitative results on DAVIS2016, SegTrack v2 and FBMS-59~(Section~\ref{sec:morevisuals}).
\end{itemize}

\section{\renaud{Justifying only using the flow between adjacent frames}}
\label{sec:usetridiagaff}

\renaud{Our formulation, in Eq.\,(1) of the main paper, only involves the optical flow between adjacent frames (forward and backward). As discussed in Section~3 of the main paper, it can be related to the tridiagonal affinity matrix $\calA$ in Eq.\,(4). We remark at the end of Section~3.3 that we could have used a denser matrix correlating more faraway frames, but that the optical flow between frames that are distant in time is less reliable.}

To validate our choice of only using the optical flow between adjacent frames, we consider here the following variant of our objective function, where we introduce warps between more distant frames (up to some constant $\Nsteps$):
\begin{equation}
\begin{array}{l}
\calL(\{\mask{p}\}_p) = \sum_p \lambda \; \CE(\maskini{p}, \mask{p}) \; + \\[0.3em]
\quad \sum_{t=1}^T\CE(\mask{p+t}, \warp{p}{p+t}(\mask{p}) ) + 
\CE(\mask{p}, \warp{p+t}{p}(\mask{p+t})) \> ,
\end{array}
\label{eq:L_multiflow}
\end{equation}

\begin{table}[ht]
\centering
\begin{tabular}{c@{}c}

  \centering
  \scalebox{1}{ % Tune this to fit the table when it is too large
    \begin{tabular}{l@{$\quad\quad$}r@{$\quad\quad$}r}
    \toprule
    \renaud{Max.\ frame distance} & \multicolumn{2}{c}{\renaud{DAVIS2016}}\\
    \renaud{for optical flow} % \nermin{No.\ flow steps} 
    & $\calJ \uparrow$  & $\calF \uparrow$ \\[0.3ex]
    \midrule
    $\Nsteps = 1$ & \textbf{76.8}  & \textbf{77.0} \\
    $\Nsteps = 2$ & 74.0  & 71.3\\
	$\Nsteps = 3$ & 67.1  & 61.8 \\
    \bottomrule \\
  \end{tabular}
  } 
\end{tabular}
\caption{\textbf{%
\renaud{Study of the distance between frames for flow consistency enforcement.}}
We consider different values of $\Nsteps$ in Eq.\,\eqref{eq:L_multiflow} and evaluate on DAVIS2016 \nermin{using our best configuration (DINO [ViT] + ARFlow + Opt)} without CRF post-processing.
The best performance is achieved for $\Nsteps \,{=}\, 1$, coinciding with the tridiagonal matrix \renaud{configuration}. 
}
\label{tab:tridiag}
\end{table}

\renaud{Table~\ref{tab:tridiag} shows that using a time horizon of a single frame is not only enough but actually better than considering the optical flow between more distant frames. In fact, 
using the flow regarding only the preceding and the succeeding frames already ties together all frames in the sequence. Additional terms with optical flows between more distant frames may actually introduce noise because of worse estimations due to larger displacements, deformations and occlusions.}

\section{Using the cross-entropy vs the dot product}
\label{sec:dotvscross}

In Section~3.4 of the main paper, we replace the \renaud{dot products}
\[
(\maskini{p})^T \mask{p} \>,\>\> 
\mask{p+1}^T \warp{p}{p+1}(\mask{p})\>\text{, and } \>\>
\mask{p}^T \warp{p+1}{p}(\mask{p+1})
\]
by cross-entropies, respectively:
\[
\CE(\maskini{p}, \mask{p})\>, \>\>
\CE(\mask{p+1}, \warp{p}{p+1}(\mask{p}))\>\text{, and }\>\> 
\CE(\mask{p}, \warp{p+1}{p}(\mask{p+1})) \>.
\]
This was motivated empirically, as we observed that the cross-entropy was providing a better performance. Table~\ref{tab:cos_vs_CE} reports the quantitative results of this experiment. 

\begin{table}[ht]
\centering
\begin{tabular}{c@{}c}

  \centering
  \scalebox{1}{ % Tune this to fit the table when it is too large
    \begin{tabular}{l@{$\quad\quad$}r@{$\quad\quad$}r}
    \toprule
    \renaud{Measurement of} & \multicolumn{2}{c}{\renaud{DAVIS2016}} \\
    mask deviation & $\calJ \uparrow$  & $\calF \uparrow$ \\
    \midrule
    Cross-entropy & \textbf{76.\renaud{8}} & \textbf{77.0} \\ % was 76.7
    Dot product & 62.1 & 60.4 \\
    \bottomrule \\
  \end{tabular}
  } 
\end{tabular}
\caption{\label{tab:cos_vs_CE} {\bf Dot product vs cross-entropy.} Using the cross-entropy between two vectors in our objective function rather than their dot product leads to a significant improvement of \textbf{+14.6\%} in $\calJ$ and \textbf{+16.6\%} in $\calF$.}
\end{table}

\section{On the \renaud{Constant Norm Constraint in Eq.\,(3)}}
\label{sec:unitnorm}

\renaud{
We estimate the second largest eigenvector of $\calW$
via a maximization problem over a vector $\bX$ under the constraint that $\|\bX\|_2$ is constant, as stated in Eq.\,(3) in the main paper. 
At the end of Section~3.4, we claim that since the $\mask{p}$ vectors remain close to the $\maskini{p}$ vectors, $\|\bX\|_2 = \sqrt{\sum_p (\|\maskini{p}\|_2)^2}$ remains approximately constant during optimization, thus satisfying the constraint in Eq.\,(3) up to a constant factor of $\sqrt{N}$.

Table~\ref{tab:unit-constraint} shows empirically that this constraint is indeed approximately met at each stage of the global optimization process.
}

\begin{table}[ht]
\centering
\begin{tabular}{ccc}

  \centering
  \scalebox{1}{
    \begin{tabular}{c@{$\quad\quad$}c@{$\quad\quad$}c}
    \toprule
    L-BFGS iteration & Theoretical            &  Actual average\\
    number    & norm \renaud{of $\maskini{p}$} &  norm \renaud{of $\maskini{p}$} \\
    \midrule
    1 & 1  & 1.003 \\
    2 & 1  & 1.001 \\
    3 & 1  & 1.007 \\
    4 & 1  & 1.022 \\
    5 & 1  & 1.022 \\
    \bottomrule \\
  \end{tabular}
  } 
\end{tabular}
\caption{\textbf{%
\renaud{Study of the constant norm constraint approximation.}}
\renaud{We study the average of $\|\maskini{p}\|_2$ over all frames of all sequences in the DAVIS2016 dataset at each iteration of our global optimization (using L-BFGS).} 
We observe that the deviations to the theoretical norm of~1 are small, which in turns means that our approach of not using any explicit constraint is valid.}
\label{tab:unit-constraint}
\end{table}

\section{Approximation in Eq.~(8)}
\label{sec:approx}

In the main paper, we assumed the following approximation:
\begin{equation}
\mask{p}^\top D_p^\inv \aff{p}\, \mask{p} \approx {\maskini{p}}^\top \, \mask{p} \> .
\label{eq:suppmat_to_prove}
\end{equation}
The derivation of this approximation is given below.

Since $W_p = D_p^\inv \aff{p}$ is a row-normalized stochastic matrix, the largest eigenvalue associated to its first eigenvector is~1. Besides, our initial mask estimate $\maskini{p}$ is computed as the second largest eigenvector of $W_p$ via Power Iteration Clustering (PIC) \cite{power-iteration}. According to \cite{meila,power-iteration}, if $K$~clusters are well-separated, then the significant eigenvalues of~$W_p$, noted $\lambda_1 \,{\geq}\, \ldots \,{\geq}\, \lambda_K$, are such that $\lambda_i/\lambda_1 \,{\approx}\, 1$ for all $i\,{\in}\, \{1,...,K\}$. Consequently, if the foreground object is well-separated from the background ($K\,{\geq}\,2$), we may assume that $\lambda_2 \,{\approx}\, \lambda_1 \,{=}\, 1$.  As $\maskini{p}$ approximates the second largest eigenvector of~$W_p$, we have:
\begin{equation}
W_p\, \maskini{p} \approx \lambda_2 \,\maskini{p} \approx \maskini{p} \> .
\end{equation}
Therefore, considering also that $\mask{p}$ deviates little from $\maskini{p}$, \ie, $\mask{p} \approx \maskini{p}$, we have:
\begin{equation}
\begin{array}{rcl}
&&\mask{p}^\top D_p^\inv \aff{p}\, \mask{p} \\
&=&\mask{p}^\top W_p\, \mask{p} \\
&\approx& \mask{p}^\top W_p\, \maskini{p}\\
&\approx& \mask{p}^\top \, \maskini{p}\\
&=& \maskini{p}^\top \, \mask{p} \> .
\end{array}
\end{equation}

\section{\vincent{Dealing with Inaccurate Optical Flow}}
\label{sec:alleviating}

Our method relies on predicted optical flows. As they can be wrong or poor quality, they may introduce noise during the computation of the initial mask estimates and the optimization. 
In order to reduce the influence of this noise, we eliminate poor quality optical flow predictions. Given a predicted flow $\flow{p}{q}{}$, we first warp frame $q$ to frame $p$. Next, we calculate the difference image between frame $p$ and the reconstructed frame $\hat{p}$. The locations with high response on the difference image correspond to wrong or poor quality optical flow predictions. We use $k$-th percentile of the difference image as a threshold value to eliminate the poor quality optical flow predictions. The locations under the calculated threshold value indicate where optical flow fails to produce the accurate flows. We exclude these optical flow predictions from both the computation of the initial mask estimates and the optimization. We experimentally set $k$ as 90-th percentile.

\section{Implementation Details}
\label{sec:implementation}

All our experiments are implemented with PyTorch~\cite{paszke-19-pytorch}. We use the L-BFGS
~\cite{byrd1995limited} with learning rate of 1 to optimize our objective function. The weight $\lambda$ in Eq.~1 is set to be $10$.

We use DINO pretrained on ImageNet as appearance features. Due to their low resolution, we upscale the initial eigenvectors to the required resolution and run the full optimization pipeline. The optimized eigenvectors can be later either thresholded or clustered with $K$-means to obtain the final masks. We choose $K$-means as it is a more universal method that does not require finding threshold parameters. The final segments are then refined using CRF, as \cite{yang2019unsupervised,yang2021dystab}.

We use the ARFlow model pretrained on the City\-Scapes \cite{Cordts2016Cityscapes} dataset in a self-supervised manner to predict optical flow. In ablation studies, we also use the RAFT model~\cite{teed-eccv20-raft} for comparison, which is trained in a supervised manner with labeled data from the Sintel dataset~\cite{Buttler2012Sintel}.

\section{Additional Failure Cases}
\label{sec:failcases}
\begin{figure}[h]
  \centering
  \begin{tabular}{c@{\;}c}
    \includegraphics[width=0.425\linewidth]{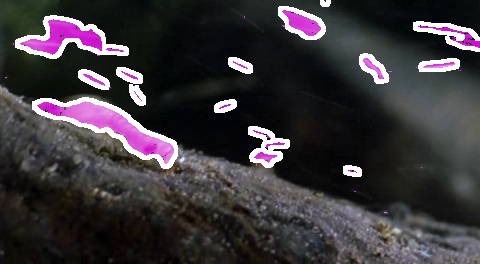} & \includegraphics[width=0.425\linewidth]{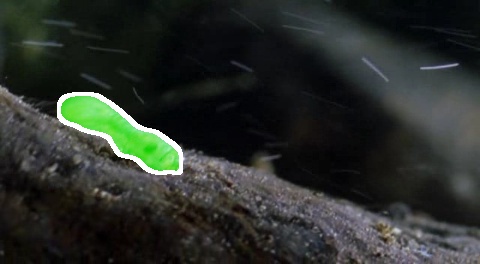} \\
    \includegraphics[width=0.425\linewidth]{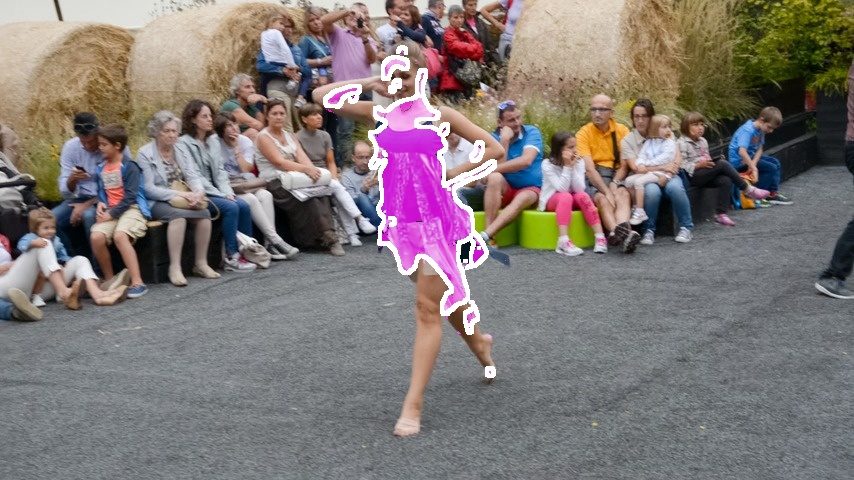} & \includegraphics[width=0.425\linewidth]{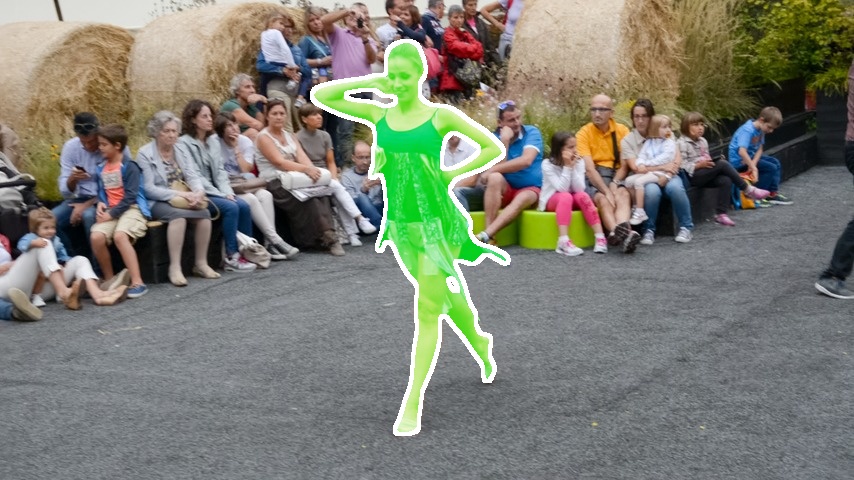} \\
    \includegraphics[width=0.425\linewidth]{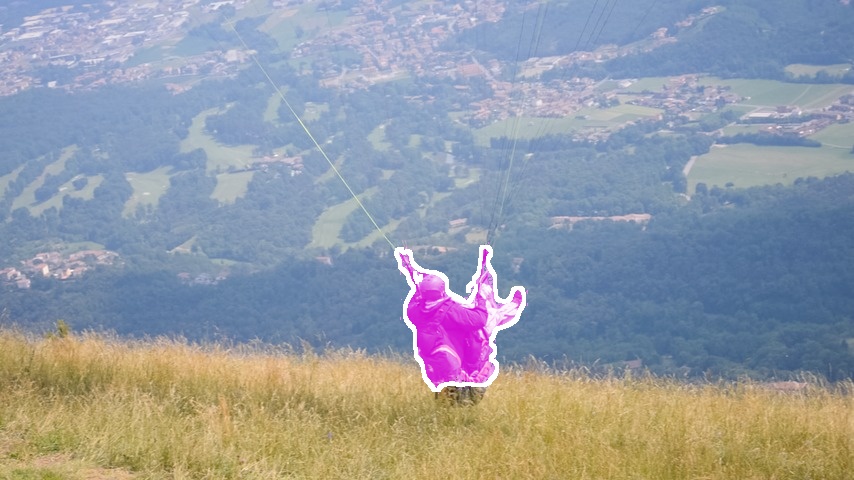} & \includegraphics[width=0.425\linewidth]{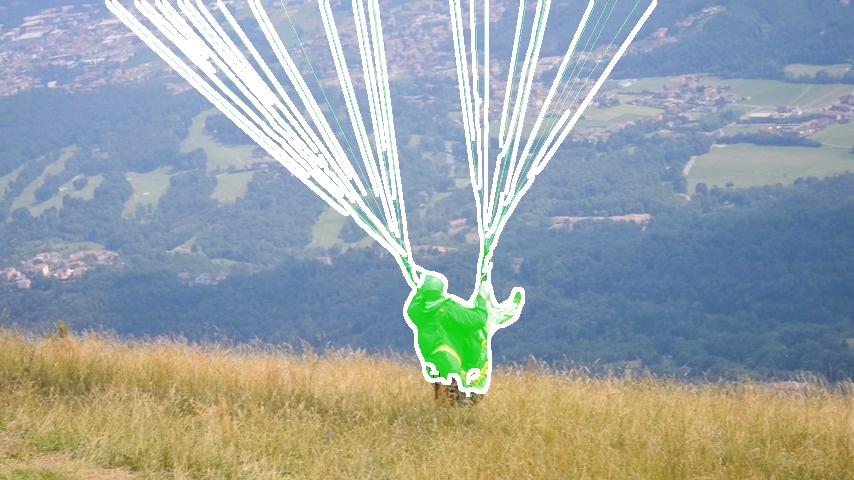} \\
    \small Ours & \small Ground Truth \\
  \end{tabular}
   \caption{\label{fig:failures_suppmat}{\bf Failure cases.} Our approach has three main failure modes: over-segmentation in scenes with multiple objects, under-segmentation, and inability to capture very fine details.}
\end{figure}

In Figure~\ref{fig:failures_suppmat}, we show more failure examples of our approach. The first row shows another example of oversegmentation, where flowing particles are being segmented as foreground. The second row  shows undersegmentation. The last row shows the inability of our approach to capture very fine details, such as the thin cables of the paraglider.

\section{More Qualitative Results}
\label{sec:morevisuals}

On the next pages (Figures~\ref{fig:supp:st1}-\ref{fig:supp:fbmslast}), we show more qualitative results of our approach, where we compare to the CIS~\cite{yang2019unsupervised} method, 
which is the third best self-supervised \renaud{video object segmentation} (VOS) method after ours, \renaud{according to Table~2 in the main paper.} \renaud{(DyStab~\cite{yang2021dystab}, which is the second best self-supervised VOS method, did not release code to rerun these experiments nor mask results).}

Compared to CIS, our method is more successful at segmenting objects as a whole and capturing finer details of object boundaries.

\newcommand{\resultsfig}[6]{
\begin{figure*}[p]
  \centering
  \begin{tabular}{c@{\;}c@{\;}c}
    \includegraphics[width=#3\linewidth]{images_suppmat/comparisons/#1/#4_cis.jpeg} & \includegraphics[width=#3\linewidth]{images_suppmat/comparisons/#1/#4_gopt.jpeg} & \includegraphics[width=#3\linewidth]{images_suppmat/comparisons/#1/#4_gt.jpeg}\\
    \hline\vspace{-3mm}\\
    \includegraphics[width=#3\linewidth]{images_suppmat/comparisons/#1/#5_cis.jpeg} & 
    \includegraphics[width=#3\linewidth]{images_suppmat/comparisons/#1/#5_gopt.jpeg} & \includegraphics[width=#3\linewidth]{images_suppmat/comparisons/#1/#5_gt.jpeg}\\
    \hline\vspace{-3mm}\\
    \includegraphics[width=#3\linewidth]{images_suppmat/comparisons/#1/#6_cis.jpeg} & 
    \includegraphics[width=#3\linewidth]{images_suppmat/comparisons/#1/#6_gopt.jpeg} & \includegraphics[width=#3\linewidth]{images_suppmat/comparisons/#1/#6_gt.jpeg}\\
    \hline\vspace{-3mm}\\
    \small CIS \cite{yang2019unsupervised} & \small Ours & \small Ground Truth 
  \end{tabular}
  \caption{Segmentation in sample frames from videos in #2.}
\end{figure*}
}

\newcommand{\resultsfigwithlabel}[7]{
\begin{figure*}[p]
  \centering
  \begin{tabular}{c@{\;}c@{\;}c}
    \includegraphics[width=#3\linewidth]{images_suppmat/comparisons/#1/#4_cis.jpeg} & \includegraphics[width=#3\linewidth]{images_suppmat/comparisons/#1/#4_gopt.jpeg} & \includegraphics[width=#3\linewidth]{images_suppmat/comparisons/#1/#4_gt.jpeg}\\
    \hline\vspace{-3mm}\\
    \includegraphics[width=#3\linewidth]{images_suppmat/comparisons/#1/#5_cis.jpeg} & 
    \includegraphics[width=#3\linewidth]{images_suppmat/comparisons/#1/#5_gopt.jpeg} & \includegraphics[width=#3\linewidth]{images_suppmat/comparisons/#1/#5_gt.jpeg}\\
    \hline\vspace{-3mm}\\
    \includegraphics[width=#3\linewidth]{images_suppmat/comparisons/#1/#6_cis.jpeg} & 
    \includegraphics[width=#3\linewidth]{images_suppmat/comparisons/#1/#6_gopt.jpeg} & \includegraphics[width=#3\linewidth]{images_suppmat/comparisons/#1/#6_gt.jpeg}\\
    \hline\vspace{-3mm}\\
    \small CIS \cite{yang2019unsupervised} & \small Ours & \small Ground Truth 
  \end{tabular}
  \caption{Segmentation in sample frames from videos in #2.}
  \label{#7}
\end{figure*}
}

\resultsfigwithlabel
{segtrackv2}{SegTrack v2}
{0.32}
{bird_of_paradise_00071}
{bmx_00001}
{drift_00069}
{fig:supp:st1}

\resultsfig
{segtrackv2}{SegTrack v2}
{0.27}
{frog_00176}
{girl_00001}
{worm_00090}
% {parachute_00021}

\resultsfig
{segtrackv2}{SegTrack v2}
{0.32}
{hummingbird_00007}
{monkey_00012}
{soldier_00022}

\resultsfig
{davis}{DAVIS 2016}
{0.32}
{blackswan_00004}
{camel_00027}
{car-roundabout_00037}

\resultsfig
{davis}{DAVIS 2016}
{0.32}
{cows_00050}
{dance-twirl_00053}
{dog_00028}

\resultsfig
{davis}{DAVIS 2016}
{0.32}
{car-shadow_00008}
{drift-chicane_00029}
{drift-straight_00014}

\resultsfig
{davis}{DAVIS 2016}
{0.32}
{goat_00017}
{horsejump-high_00008}
{libby_00012}

\resultsfig
{davis}{DAVIS 2016}
{0.32}
{motocross-jump_00011}
{parkour_00005}
{soapbox_00069}

\resultsfig
{fbms59}{FBMS-59}
{0.32}
{camel01_00000}
{cars1_00000}
{cats01_00059}

\resultsfig
{fbms59}{FBMS-59}
{0.32}
{rabbits02_00000}
{cats03_00079}
{dogs02_00259}

\resultsfig
{fbms59}{FBMS-59}
{0.32}
{farm01_00159}
{giraffes01_00000}
{tennis_00009}

\resultsfig
{fbms59}{FBMS-59}
{0.32}
{goats01_00159}
{horses02_00079}
{lion01_00155}

\resultsfigwithlabel
{fbms59}{FBMS-59}
{0.32}
{people03_00119}
{rabbits03_00000}
{rabbits03_00000}
{fig:supp:fbmslast}

\section{Use of Existing Datasets and Codes}
\label{sec:assets}

For the experiments, we used several datasets that are freely available for research purpose:
\begin{itemize}

    \item DAVIS 2016\,\footnote{\url{https://davischallenge.org}}~\cite{perazzi-cvpr16-abenchmarkdataset} is under license CC BY-NC~4.0.

    \item SegTrack-v2\,\footnote{\url{https://web.engr.oregonstate.edu/~lif/SegTrack2/dataset.html}}~\cite{li-iccv13-videosegmentation} is under a custom non-commercial, research-only license, courtesy of Georgia Institute of Technology,

    \item FBMS-59\,\footnote{\url{https://lmb.informatik.uni-freiburg.de/resources/datasets/moseg.en.html}}~\cite{Brox14,brox2010object} is under a custom non-commercial, research-only license, courtesy of University of Freiburg.
\end{itemize}

To compute appearance and flow, we experimented with the following methods, whose code is freely available for research purpose:
\begin{itemize}

    \item DINO\footnote{\url{https://github.com/facebookresearch/dino}} \cite{caron2021emerging} is under the Apache License 2.0.
    
    \item MoCov2\footnote{\url{https://github.com/facebookresearch/moco}} \cite{chen2020mocov2} is under the CC BY-NC~4.0 license.
    
    \item ARFlow\footnote{\url{https://github.com/lliuz/ARFlow}} \cite{liu2020learning} is under the MIT License.

    \item RAFT\footnote{\url{https://github.com/princeton-vl/RAFT}} \cite{teed-eccv20-raft} is under the BSD 3-Clause License.

\end{itemize}

\section{Societal Impact}
\label{sec:socialimpact}

We believe that our approach for the self-supervised discovery and segmentation of objects in videos has \textbf{very little potential for malicious uses} (including disinformation, surveillance, invasion of privacy, endangering security), in any case not more, \eg, than the hundreds of previously published methods on supervised object detection and segmentation. Moreover, we are not bound nor promoting any dataset that would lead to unfairness in any sense.

Besides, the use of our method has a \textbf{very little environmental impact} as there is no training phase and as the optimization is relatively fast and in the same order of magnitude as other approaches.

\end{document}